# A Comprehensive Review of Knowledge Distillation in Computer Vision

**Gousia Habib*[1], Tausifa jan Saleem[1], Sheikh Musa Kaleem[2], Tufail Rouf [2], Brejesh Lall[1]**

1. *Bharti School of Telecommunication Technology and management, Indian Institute of Technology New Delhi, India*

2. *National Institute of Technology Srinagar*

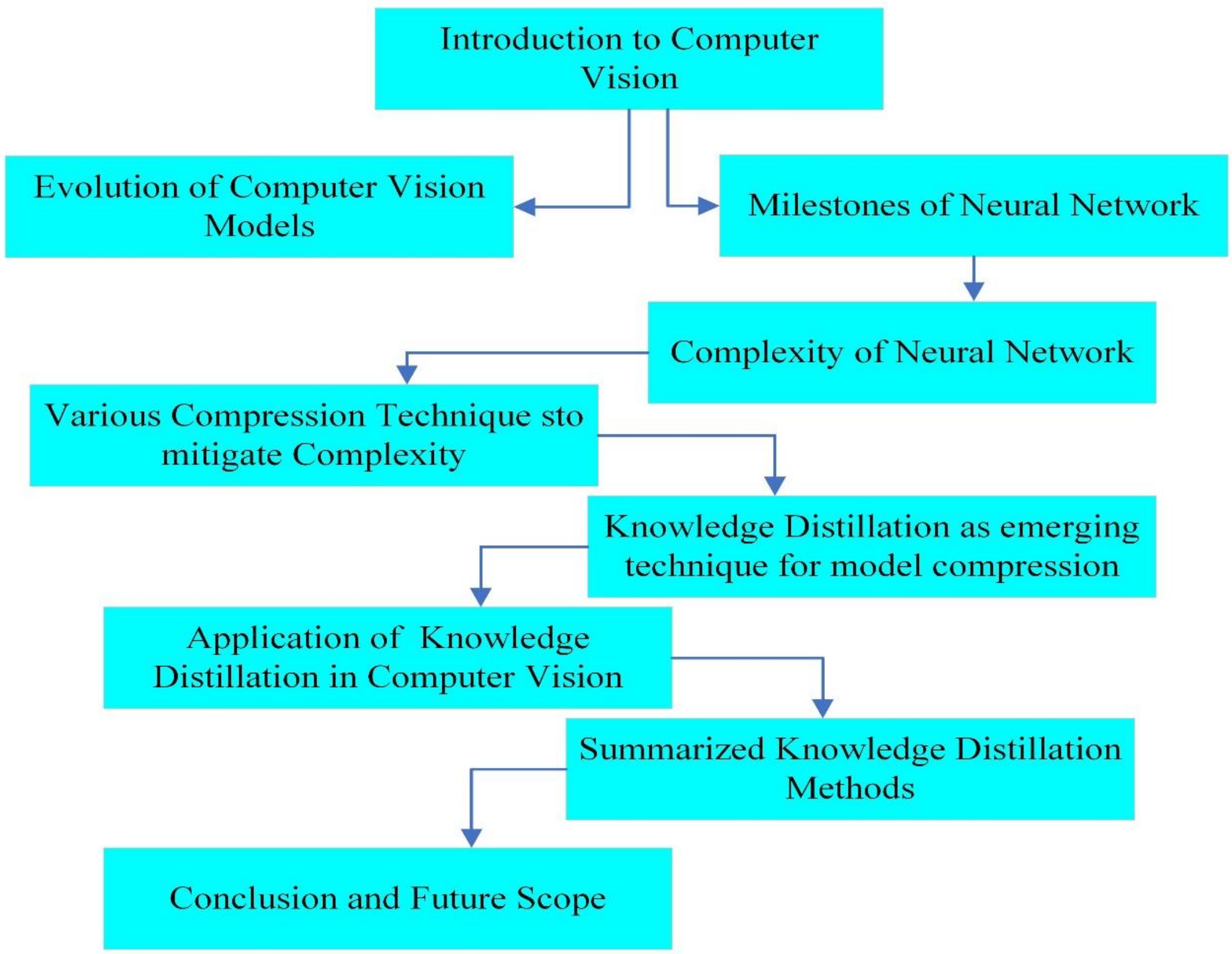

**Taxonomy Highlighting Key Themes of paper**

## ABSTRACT

*Deep learning techniques have been demonstrated to surpass preceding cutting-edge machine learning techniques in recent years, with computer vision being one of the most prominent examples. However, deep learning models suffer from significant drawbacks when deployed in resource-constrained environments due to their large model size and high complexity. Knowledge Distillation is one of the prominent solutions to overcome this challenge. This review paper examines the current state of research on knowledge distillation, a technique for compressing complex models into smaller and simpler ones. The paper provides an overview of the major principles and techniques associated with knowledge distillation and reviews the applications of knowledge distillation in the domain of computer vision. The review focuses on the benefits of knowledge distillation, as well as the problems that must be overcome to improve its effectiveness.*

## 1) INTRODUCTION

Computer vision is a subfield of computer science and artificial intelligence (AI) that focuses on teaching robots to interpret and grasp visual information in their environment. It entails the creation of algorithms and procedures that enable computers to analyze, process, and interpret digital images and videos to extract meaningful information. Computer vision is an interdisciplinary field that draws from a variety of disciplines, including mathematics, physics, statistics, computer science, and engineering. Object detection and recognition [1], image and video processing [2], autonomous cars [3], robotics [4], medical imaging [5], and security and surveillance systems [6] are some of the primary applications of computer vision. The growing availability of big datasets, powerful computing resources, and effective training techniques have accelerated the development of deep learning-based computer vision models. In certain tasks, such as object detection [8-9] in natural images [10-11], these models [12] can currently outperform human specialists.

In the field of computer vision, notable and substantial advances have been made in recent years, with deep learning techniques playing a pivotal role in achieving breakthroughs in various vision tasks (Figure 1). Deep learning comprises several architectures, each with its own set of properties. These are shown in the table below. The growing complexity and resource demands of deep learning models pose challenges for their deployment in resource-constrained environments or real-time applications. Knowledge distillation has emerged as a promising technique to address these challenges by compressing large-scale models into smaller and more efficient counterparts without significantly compromising performance. Throughout this paper, we survey a wide range of computer vision applications that have leveraged knowledge distillation. We investigate how knowledge distillation has been employed in tasks such as image classification, object detection, semantic segmentation, image generation, and more. We explore the specific techniques and architectures used in these applications, highlighting the benefits and trade-offs associated with knowledge distillation. By examining the latest advancements in the field and drawing insights from a diverse set of applications, this review paper aims to provide researchers and practitioners with a comprehensive understanding of the application of knowledge distillation in computer vision. We shed light on the potential benefits, limitations, and future directions of this technique, paving the way for further advancements in the efficient utilization of deep learning models in computer vision tasks. Fig. 1 shows the growth in computer vision, image analysis, machine learning, image processing, and pattern recognition in the past 20 years.

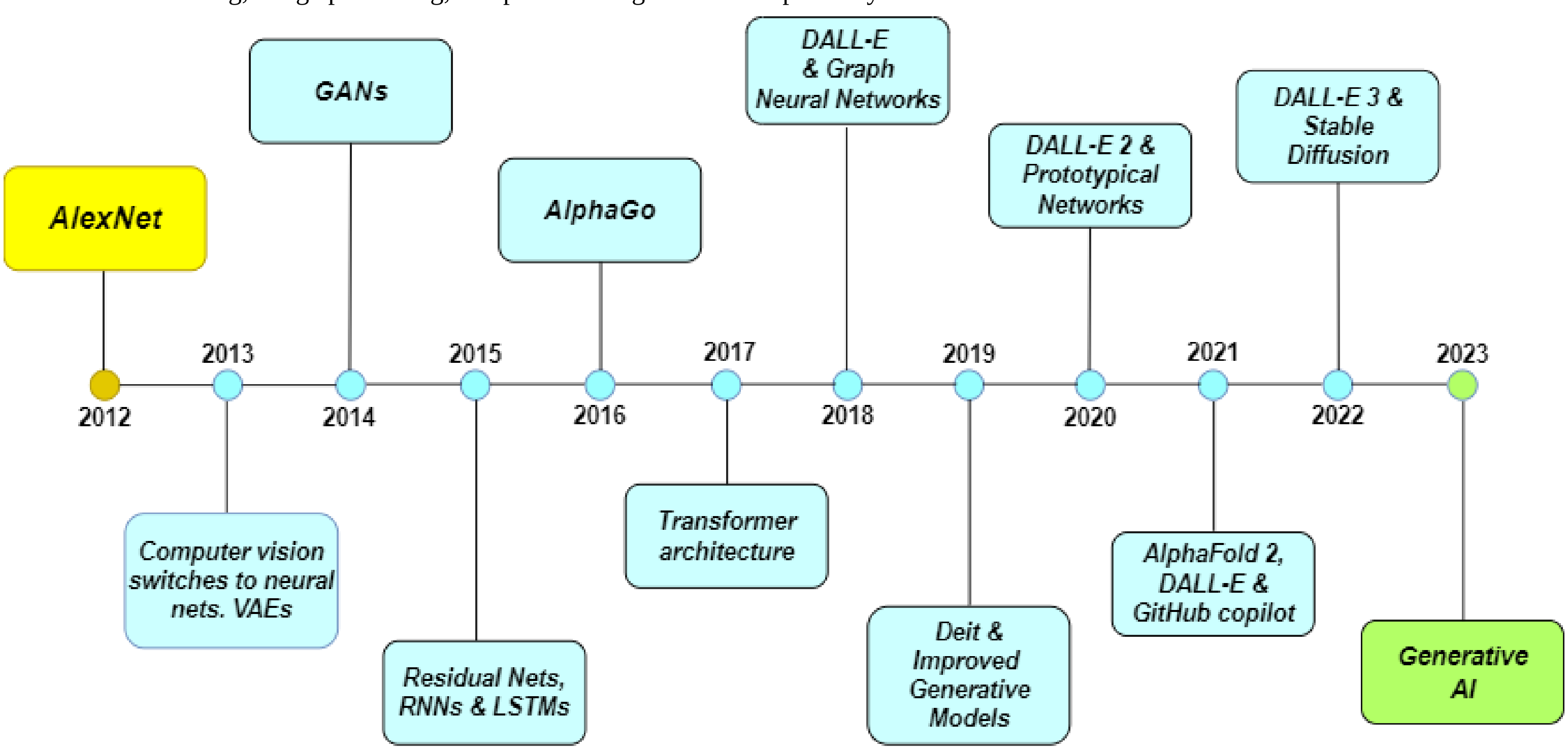

Fig 1: Computer Vision milestones

In recent years, key milestones and advancements in computer vision have revolutionized the way machines interpret and process visual information. Table 1 outlines several major milestones in computer vision, highlighting breakthroughs that have opened the path for the use of knowledge distillation.

Table 1: Some significant milestones in the history of computer vision

| | | | |
|---|---|---|---|
| **1966** | First computer vision system | The first computer vision system was developed by Larry Roberts at MIT in 1966. The system was able to recognize simple objects in images and could distinguish between different shapes such as circles, squares, and triangles. | **[13]** |
| **1973** | Generalized Cylinder | The "Generalized Cylinder" representation was introduced by Andrew Witkin and Michael Klass and was used to model 3D objects in computer vision. | **[14]** |
| **1974** | Marr Vision Model | David Marr developed a theoretical framework for understanding visual perception, known as the "Marr Vision Model". This model described the process of perception as a sequence of computational stages, starting with the raw visual input and ending with a 3D model of the scene. | **[15]** |
| **1980s** | Computer vision used for business | The first successful commercial application of computer vision was developed by the Japanese company Omron, which created a system capable of inspecting IC chips for defects. | **[16]** |
| **1986** | Scale Invariant Feature Transform | David Lowe created the "Scale Invariant Feature Transform" algorithm, which could find and match features in images that are resistant to changes in size, rotation, and lighting. | **[17]** |
| **1991** | Viola-Jones | Paul Viola and Michael Jones created the "Viola-Jones" algorithm, which uses machine learning and Haar-like features to recognizes faces in real-time. | **[18]** |
| **1999** | Lucas-Kanade | Bruce Lucas and Takeo Kanade developed the "Lucas-Kanade" algorithm for optical flow estimation, which was used to detect movement in videos. | **[19]** |
| **2001** | Bag of Visual Words | Svetlana Lazebnik and Cordelia Schmid developed the "Bag of Visual Words" (BoVW) concept, which divided images based on their visual elements. | **[20]** |
| **2012** | AlexNet | Alex Krizhevsky and colleagues created the "AlexNet" neural network design, which significantly improved image categorization accuracy | **[21]** |
| **2014** | DeepFace | The "DeepFace" algorithm was developed by researchers at Facebook, which could recognize faces with high accuracy even in challenging conditions. | **[22]** |
| **2015** | Generative Adversarial Networks | Ian Goodfellow and his colleagues proposed "Generative Adversarial Networks" (GANs), which allowed for the creation of realistic images and films from noise. | **[23]** |
| **2018** | Capsule Networks | The "Capsule Networks" architecture was proposed by Geoffrey Hinton and Sara Sabour, which aimed to overcome the limitations of traditional neural networks in modelling spatial relationships between objects in images. | **[24]** |
| **2020** | DALL-E | The "DALL·E" system was introduced by OpenAI, which could generate highly realistic images from textual descriptions, demonstrating the potential of artificial intelligence to transform the field of computer vision. | **[26]** |
| **2022** | **DALL·E 2** | DALL-E 2 is an OpenAI generative neural network that can generate realistic visuals from text descriptions. It is a follow-up to the original DALL-E model, which debuted in 2020. | **[27]** |

### 1.1.1 Neural Networks (NN)

Neural networks are a fundamental concept in artificial intelligence and machine learning, inspired by the structure and functioning of the human brain. Comprising interconnected nodes or "neurons," organized into layers, neural networks process information to perform tasks such as pattern recognition and decision-making. Input data is fed into the input layer, processed through hidden layers using weighted connections, and the final output is generated in the output layer. During training, the network adjusts the weights of connections through a process called backpropagation, minimizing the difference between predicted and actual outcomes. Neural networks learn to generalise and produce precise predictions on fresh, unobserved data. Neural networks come in several varieties, such as feedforward, recurrent, and convolutional networks, and are each intended to perform a particular function, like image recognition, sequential data analysis, etc. [28]

CNNs are a prominent deep learning approach as it pertains to computer vision. Among their many applications, CNNs are particularly suitable for image and video analysis tasks such as image classification, object identification, semantic segmentation, and others. CNNs are used in computer vision to extract characteristics from images or videos that may then be used to generate predictions or perform actions based on the content of those images or videos. CNNs do this by applying filters to the input data via convolutional layers, effectively scanning the image or video for patterns or features important to the task at hand. Overall, CNNs have transformed the area of computer vision by allowing machines to execute jobs that were previously only performed by humans. They are currently an essential component of many computer vision systems and remain an active area of study and development. CNN is composed of layers such as convolutional layers, pooling layers, and fully linked layers. A CNN extracts information from an input image by processing it through a series of convolutional layers. The output of each convolutional layer is then passed via a pooling layer, which down samples the output to make it smaller. To classify the image based on the extracted features, the output of the last pooling layer is flattened and passed through one or multiple fully connected layers [29].

### 1.1.2 Neural Network Milestones

Numerous prominent Neural Network (NN) designs have made major contributions to the field of computer vision. The table below (Table 2) summarizes all NN architectures when they were invented, and their accuracy:

Table 2: Summary of NN architectures

| Name | Date | Author | Data Set | (Top-5) Error Rate | Number of parameters (approximately) | Number of FLOPS | Reference |
|---|---|---|---|---|---|---|---|
| **Perceptrons** | 1957 | Frank Rosenblatt | Small Linear Data set | 5% | 12 - 1000 | 1000 | **[30]** |
| **Multilayer Perceptrons** | 1970s | Marvin Minsky and Seymour Papert | Small (hundreds-thousands) | 5-10% (linear), 30%+ (non-linear) | Tens of thousands to millions | Tens of thousands to millions | **[31]** |
| **Backpropagation** | 1986 | Geoffrey Hinton, David Rumelhart, and Ronald Williams | Millions or even billions of data points | 0.1% or lower | Millions or billions (e.g., ResNet-50: 25.6 million, GPT-3: 175 billion) | Millions or billions per training iteration (e.g., ResNet-50: 4 billion per image) | **[32]** |
| **The Long Short-Term Memory (LSTM) architecture** | 1997 | Sepp Hochreiter and Jürgen Schmidhuber | Large (millions-billions) | 1-5% | Millions to billions | Millions to billions | **[33]** |
| **Gated Recurrent Units (GRU)** | **2014** | **Cho** | **Large (millions-billions)** | **1-5%** | **Millions to billions** | **Millions to billions** | **[34]** |
| **LeNet** | **1998** | **Yann LeCun** | **MNIST** | **0.95%** | **60,000** | **189K** | **[35]** |
| **AlexNet** | 2012 | Alex Krizhevsky | ILSVRC 2012 | 15.3% | 60 million | 720M | **[36]** |
| **VGG** | 2014 | Karen Simonyan and Andrew Zisserman from the University of Oxford. | ILSVRC 2014 | 7.3%, | (VGG16) 138 million (VGG19) 143 million | 15.5 billion | **[37]** |
| **GoogleNet** | 2014 | Christian Szegedy, Wei Liu, and colleagues | ILSVRC 2014 | 6.7% | 6.8 million | 1.5 billion | **[38]** |
| **ResNet** | 2015 | Xiangyu Zhang, Shaoqing Ren, Jian Sun, and Kaiming He from Microsoft Research. | ILSVRC 2015 | 3.57% | 25.6 million | 4.1 billion | **[39]** |
| **Inception v2, v3, and v4** | 2014-2016 | Christian Szegedy, Vincent Vanhoucke, and colleagues from Google Research | ILSVRC | (Inception v 2) 3.16%, (Inception v3) 3.08%, (Inception v4) 2.76% | (Inception v2) 11.2 million (Inception v3) 23.8 million, (Inception v4) 42.7 million | Inception-v2: 11.2 billion Inception-v3: 5.7 billion Inception-v4: | **[40]** |
| | | | | | | 12.0 billion | |
| **Xception** | 2016 | François Chollet, the creator of the Keras deep-learning library | ImageNet | 3.5% | 22.9 million | 8.9 billion | **[41]** |
| **MobileNet** | 2017 | Andrew G. Howard, Menglong Zhu, and colleagues from Google Research. | ImageNet | 10.5% | 4.2 million | 569 million | **[42]** |
| **DenseNet** | 2017 | Gao Huang, Zhuang Liu, Laurens van der Maaten, and Kilian Q. Weinberger | ImageNet | 7.8%. | 7 million | 2.8 billion | **[43]** |
| **EfficientNet** | 2019 | Mingxing Tan and Quoc V. Le from Google Research. | ImageNet | 3.3%, | 5.3 million | Efficient Net- B0 has approximately 0.39 billion flops, while Efficient Net- B1 has around 0.71 billion flops. | **[44]** |
| **Vision Transformer (ViT)** | 2020 | Alexey Dosovitskiy, Lucas Beyer, Alexander Kolesnikov, and colleagues. | ImageNet | 2.6%. | 86 million | 4.0 billion | **[45]** |
| **RegNet** | 2020 | Ilija Radosavovic, Raj Prateek, and colleagues from Facebook AI Research (FAIR). | ImageNet | 5.0% | 22 million | 4.2 billion | **[46]** |
| **ResNeSt** | **2020** | **Hang Zhang, Chongruo Wu, and colleagues.** | **ImageNet** | **2.1%** | **27 million** | **4.1 billion** | **[47]** |
| **RepVGG** | **2021** | **Xiaohan Ding, Xiangyu Zhang, and colleagues.** | **ImageNet** | **6.3%** | **12.6 million** | **4.4 billion** | **[48]** |
| **DALL-E** | 2021 | The specific author names for the DALL-E model have not been publicly disclosed. | | It is important to note that DALL-E is not typically evaluated based on a traditional acc uracy metric. Instead, its performance is assessed based on the quality and fidelity of the generated images, as well as the ability to align with the given textual prompts. | 12 billion | 6 billion | **[26]** |
| **CLIP (Contrastiv e Language-Image Pre-Training)** | 2021 | Alec Radford, Ilya Sutskever, Jong Wook Kim, Gretchen Krueger, and Sandhini Agarwal. | ImageNet | CLIP scored a top-1 accuracy of 86.4% | 33 million | | **[49]** |
| **GPT 4** | **2023** | **Olivier Caelen and Marie-Alice Blete** | **Data Grid** | **GPT-4 produced 96% accuracy when predicting category and 89% accuracy when predicting both category and sub-category.** | **1 trillion** | **22 quintillio n** | **[50]** |

### 1.1.3 Complexity of the Neural Network

Neural networks (NNs) have revolutionized countless fields, from computer vision and natural language processing to healthcare and finance. Their ability to learn complex patterns and adapt to diverse data has yielded remarkable results. However, their undeniable power comes with a caveat: complexity. Which can be best explained in terms of following parameters: [28]

- **High Parameter Count**: NNs, especially deep ones, have a vast number of parameters, often reaching into the millions or even billions. This translates to increased computational resources for training and inference, limiting their deployment on resource-constrained devices.

- **Interpretability Challenge**: NNs, for all their effectiveness, can be opaque "black boxes." Understanding how they arrive at their decisions can be challenging, leading to concerns about bias and trustworthiness.

- **Data Dependency**: NNs heavily rely on large amounts of high-quality data for effective training. This can be a barrier for applications where data is scarce or expensive to acquire.

However neural networks (NN) are complex models, with intricate connections between artificial neurons. Convolutional Neural Networks (CNNs) simplify this complexity, especially in visual tasks. CNNs use shared weights to efficiently learn hierarchical features from images, addressing challenges like spatial hierarchies and large datasets. This makes CNNs particularly effective in tasks such as image recognition and computer vision.

**Convolutional Neural Networks (CNNs)**: CNNs have achieved remarkable success in computer vision tasks like object recognition, image classification, and scene understanding. Their ability to exploit the spatial structure of images allows them to extract intricate features and achieve state-of-the-art accuracy. A summary of the complexity of CNN is given in table 3 below. Table 3 presents the complexity of a CNN layer-wise.

Table 3: Complexity of CNN

| **Parameter** | **Convolutional Layer** | **Pooling Layer** | **Fully connected Layer** |
|---|---|---|---|
| **Illustration** | 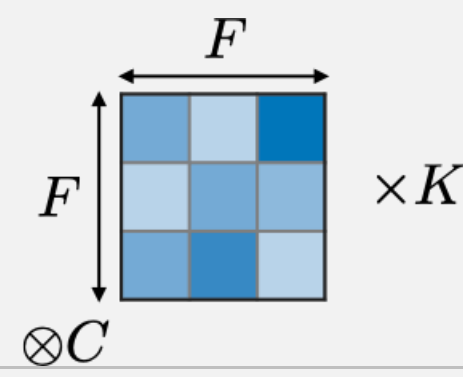 | 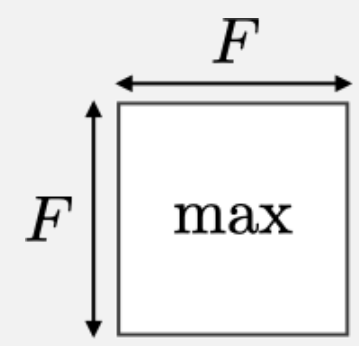 | 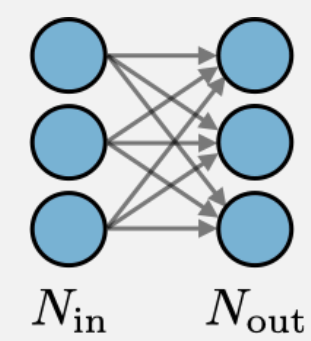 |
| **Input Size** | $(I \times I \times C)$<br>(Where I*I denotes the height and width of the image and C denotes the number of channels) | $(I \times I \times C)$ | $\boldsymbol{N_{in}}$<br>**(Nin denotes the number of input neurons)** |
| **Output Size** | $(O \times O \times K)$<br>(O*O represents the height and width of the output image with k number of filters) | $(O \times O \times C)$ | $\boldsymbol{N_{out}}$<br>**(Nout denotes the number of output neurons)** |
| **Number of Parameters** | $(F \times F \times C + 1) \cdot K$<br><br>(F*F) denotes kernel size or filter size and K denotes the number of filters for performing convolution operation. | 0 | $\boldsymbol{(N_{in} + 1) \times N_{out}}$ |
| **Remarks** | • **Each filter has a single bias parameter.**<br><br>• **S < F is typically used. S denotes the stride.**<br><br>• **The standard option for K is 2C** | • **Pooling operation carried out in a channel.**<br><br>• **Most of the time, S=F.** | • **Flattening of input.**<br><br>• **One bias parameter per neuron.**<br><br>• **No structural restrictions on the number of FC neurons.** |

Each convolutional layer filter has a single bias parameter. S < F is commonly utilized, while 2C is the conventional choice for K. In the pooling layer mostly S=F. In a fully connected layer, the input is flattened. Each neuron has one bias parameter and there are no structural constraints on the number of fully connected neurons. CNNs are sophisticated deep-learning models that are widely utilized in computer vision. CNN, in contrast, may be computationally complicated and resource-intensive, bringing these models to resource-constrained scenarios, like mobile devices, where their computational requirements may be difficult, poses a challenge. or embedded systems. CNNs' complexity and computation arise from their huge number of parameters as well as the expensive calculations required for convolutions and pooling processes. These issues can result in greater memory needs, longer inference times, and substantial energy consumption. Several ways have been proposed to make CNNs practical in resource-constrained contexts. Model compression is one strategy that seeks to minimize the size and computational needs of CNNs while preserving their performance. Network Pruning [8], Weight Quantization [9], Weight multiplexing [51], Parameter Sharing [11], and Knowledge Distillation are examples of techniques used to limit the number of parameters or the precision of weights and activations. Network architecture design is another way. Compact network topologies, such as MobileNet and Squeeze Net, are purpose-built to have fewer parameters and reduced computing requirements, making them more suited to resource-constrained applications.

To minimize the complexity and size of a CNN, network pruning includes deleting unimportant connections, neurons, or filters. The network gets more efficient in terms of memory utilization, computation, and inference speed by pruning redundant or less essential components. Pruning can be done based on a variety of parameters, including weight magnitude, activation response, and sensitivity analysis. It aids in the reduction of overfitting, the improvement of generalization, and the optimization of resource utilization. Weight multiplexing is a method that decreases the number of distinct weights in a CNN. It is also known as weight sharing or parameter sharing. Weight multiplexing allows several connections to share the same weight value rather than allocating distinct weights to each connection. This minimizes the network's memory footprint and enables for more efficient storage and computing. Convolutional kernels with shared weights and factorized convolutional layers are two common weight multiplexing approaches.

The goal of weight quantization is to lower the accuracy of weight values in a CNN. Weight quantization decreases the number of bits needed to represent weights instead of utilizing full precision (e.g., 32-bit floating-point integers). For example, weights might be quantized to 8-bit integers or even smaller. By utilizing specialized hardware for low-precision arithmetic, this strategy decreases memory needs and potentially speeds up calculations. Although quantization reduces accuracy, it is frequently compensated for via procedures such as retraining or calibration. However, convolutional Neural Networks (CNNs) are complex in visual tasks, requiring layered architectures and shared weights. Vision Transformer (ViT) simplifies this complexity by treating images as sequences of patches, using a transformer architecture. This eliminates grid-like structures, allowing ViT to efficiently capture global dependencies through self-attention, overcoming the complexities associated with CNNs.

**Vision Transformers (ViTs):** ViTs, inspired by the success of transformers in natural language processing, have emerged as a powerful alternative to CNNs in computer vision. They excel at capturing long-range dependencies and contextual relationships, leading to improved performance on tasks like image generation and action recognition. Vision transformers have demonstrated remarkable performance in computer vision tasks, particularly image classification, by leveraging self-attention mechanisms. However, a notable challenge associated with Vision Transformers is their quadratic time complexity, which stems from the self-attention mechanism's pairwise computation between all tokens in the input sequence. As the length of the input sequence increases, such as in the case of high-resolution images, the computational cost grows quadratically, leading to increased training and inference times. This poses scalability concerns, particularly for large datasets and high-resolution images, hindering the widespread adoption of Vision Transformers in resource-constrained environments. Efforts to address this challenge involve exploring alternative attention mechanisms, model parallelism, and efficient attention approximation techniques to alleviate the computational burden while preserving the expressive power of Vision Transformers in image understanding tasks. To mitigate the computational challenges posed by the quadratic time complexity of Vision Transformers (ViTs) due to the attention mechanism, researchers have explored various approaches, including pruning and quantization.[52]

**Pruning:** Pruning involves the removal of redundant or less critical connections in the neural network. In the context of ViTs, this technique can be applied to the attention mechanism to reduce the number of pairwise computations. By identifying and eliminating less influential attention weights or attention heads during training or post-training, pruning helps decrease the overall model size and computational requirements. Pruning is an effective strategy for enhancing the efficiency of Vision Transformers while maintaining competitive performance.

**Quantization:** Quantization addresses the computational challenges by reducing the precision of model weights. Instead of representing weights with high precision (e.g., 32-bit floating-point numbers), quantization allows weights to be expressed with a lower bit precision (e.g., 8-bit integers). This reduction in precision not only decreases the memory footprint of the model but also accelerates the computation during training and inference. Quantization is a widely used technique for optimizing neural network efficiency, including Vision Transformers, without significantly compromising their predictive capabilities.

However, Vision Transformer (ViT) complexity arises from attention mechanisms and sequence processing in image data. Knowledge Distillation simplifies ViT by transferring insights from a larger model, allowing for a more compact representation. This process enhances efficiency, making ViT more suitable for scenarios with limited computational resources or real-time requirements.

Knowledge distillation has emerged as effective technique for making CNNs practical in such situations. It involves teaching a smaller, more compact model (the student) to behave like a larger, more accurate model (the teacher). The student model can achieve equivalent performance with much fewer parameters and lower processing needs by passing information from the teacher to the student. Knowledge distillation provides a viable approach by balancing model size, performance, and computational efficiency. It enables resource-constrained devices to benefit from larger model's knowledge without suffering the same computing load. Furthermore, knowledge distillation enables transfer learning, in which a pre-trained teacher model is used to train a student model on a given task or dataset, enhancing efficiency even further in resource-constrained circumstances.

## 1.2 INTRODUCTION TO KNOWLEDGE DISTILLATION

A powerful machine learning approach called knowledge distillation enables the transfer of knowledge from a large and complex model to a more compact and computationally efficient model. The idea of knowledge distillation was first introduced by Bucila et al. [12] in 2006, and since then it has gained significant attention from researchers and practitioners in the field of machine learning.

Training a smaller model, known as a student model, to do the same tasks as the bigger model, known as a teacher model, is the fundamental aim of knowledge distillation. To achieve cutting-edge performance, the teacher model is often trained on a specific task and dataset. On the other hand, the student model is smaller and more computationally effective, making it suitable for deployment on devices with constrained resources. Transferring knowledge from the teacher model to the student model allows us to improve the performance of the student model without appreciably raising its complexity. The student model is trained using the same task and dataset as the teacher model throughout the knowledge distillation process (figure 2). The student model is trained according to the teacher model's predictions, which provide the student model additional knowledge and help it perform better. Knowledge transfer techniques include soft target training, attention transfer, and feature mimicry [52].

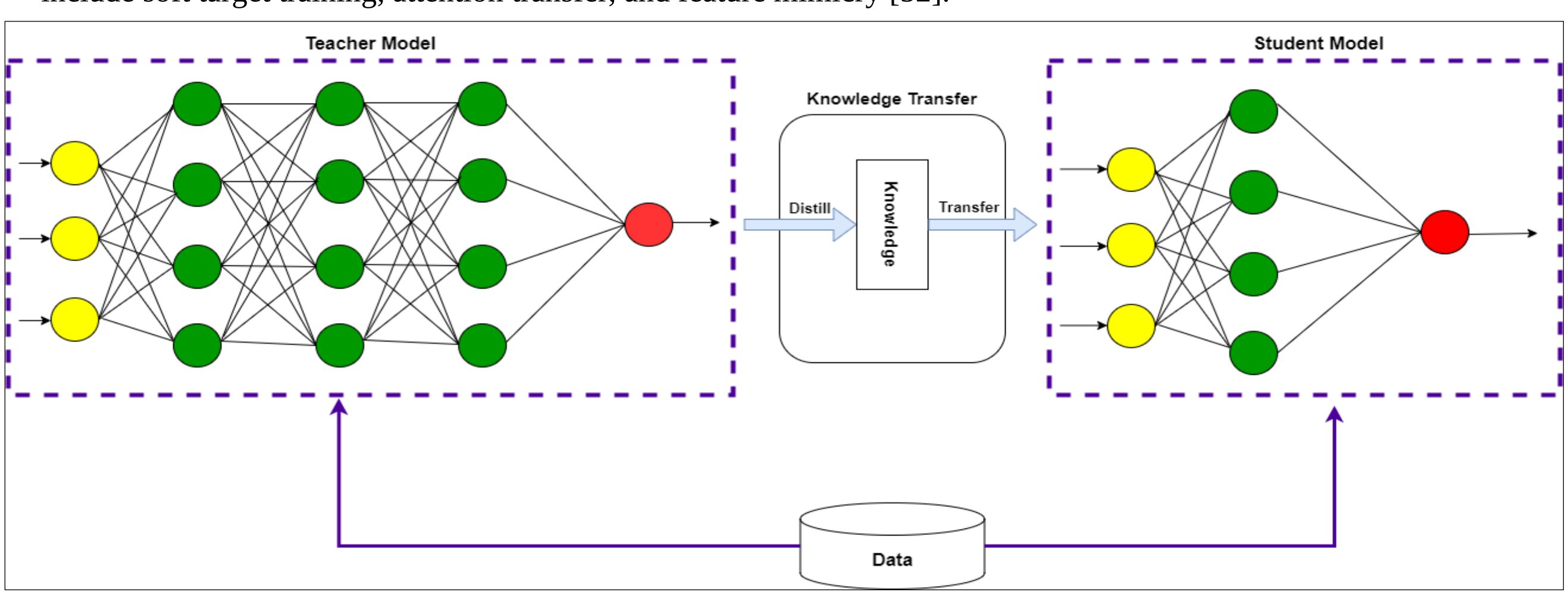

Fig 2: The generic teacher-student framework for knowledge distillation [50]

In the distillation loss, the lightweight student model duplicates the output produced by the teacher model by using the loss function. The weighted total contains the cross-entropy loss between the student's output and the real labels, as well as the cross-entropy loss between the student's output and the teacher's output (both after temperature scaling).

$$L_{distil} = \alpha * T^2 * cross\ entropy\ (y_{student,\ yteacher}/T) + (1 - \alpha) * cross\ entropy\ (y_{student}, t) \qquad (1)$$

where $y_{student}$ is the student model output, $y_{teacher}$ is the teacher model output, t is the true labels, $\alpha$ is a hyperparameter that regulates the weight given to each term, and T is the temperature parameter used for temperature scaling. The use of temperature scaling to soften soft logits produced by the softmax function is another important feature of knowledge distillation. This is accomplished by dividing the logits by a temperature parameter T, followed by the softmax function. The temperature parameter governs the probability distribution's "softness," with higher temperatures resulting in softer distributions. Furthermore, the distillation loss contains a hyperparameter called alpha, which regulates the weight assigned to each phrase. In the distillation loss function, the value of alpha defines the relative relevance of the two cross-entropy losses. In practice, alpha is usually set between 0.5 and 1, with a greater value suggesting that the output of the teacher model is given more weight. By minimizing the distillation loss, the student model learns to produce outputs that are not only accurate but also similar to those of the teacher model. This can be beneficial in scenarios where computational resources are limited and a smaller model is preferred, but the accuracy of the larger model is still desired.

## 1.3 Types of Knowledge Transfer

The following presents the different types of knowledge transfers;

*1.3.1. Response-based knowledge transfer*

The teacher model's predictions are immediately applied to the student model in this type of knowledge distillation. By training the student model to predict the same class labels or a soft target based on the output probabilities of the teacher model, response-based knowledge transfer attempts to train the student model to predict the same class labels or a soft target. This kind of knowledge transfer is frequently used when the teacher model is a deep neural network and the student model is a smaller neural network. Despite being simpler and having fewer parameters than the teacher model, the student model may benefit from the teacher model's behavior by employing predictions as targets.

*1.3.2. Feature-based knowledge transfer*

It is a form of knowledge distillation in which the intermediate feature representations that the teacher model has learned are taught to the student model. Using feature-based knowledge transfer, the student model is taught to mimic the teacher model's feature representations, which can capture implicit data knowledge. When a teacher model is a deep neural network and the student model is a shallower neural network, this type of information transfer is widely used. The student model may benefit from the teacher model's capacity to spot complex patterns in data by using feature representations from the teacher model as benchmarks to learn from. Feature-based knowledge transfer and response-based knowledge transfer may both be used in the attention transfer strategy, where the student model is trained to match the attention maps of the teacher model. The student model may have access to a range of knowledge sources by combining different information transfer strategies, enhancing performance and generalization. Last but not least, feature-based knowledge transfer can be a useful technique for knowledge distillation since it allows the student model to gain understanding from the teacher model's implicit comprehension of the data rather than just its predictions for the outcome.

*1.3.3. Relation-based knowledge transfer*

It is a sort of knowledge distillation in which the acquired relationships between the teacher model's inputs and outputs are transferred to the student model. The purpose of relation-based knowledge transfer is to teach the student model the same relationships between inputs and outputs as the teacher model, which can capture implicit data knowledge. This kind of knowledge transfer is typically used when the teacher model is a decision tree or a rule-based model and the student model is a simpler decision tree or a collection of rules. Even though it is less complex and has fewer rules than the teacher model, the student model can benefit from the teacher model's decision-making process by transmitting the understood links between the inputs and outcomes. Relationship-based knowledge transfer can be paired with feature-based information transfer, as seen in the neural additive model's technique, in which the student model is taught to imitate the additive structure of the teacher model. The student model can benefit from many sources of knowledge by combining distinct methods of knowledge transfer, resulting in enhanced performance and generalization.

Overall, relation-based information transfer can be an effective method for knowledge distillation, particularly when the teacher model is a decision tree or a rule-based model. The student model can capture the teacher model's decision-making process by transferring the learned relationships between the inputs and outputs, resulting in enhanced performance and interpretability [54]. The different types of knowledge transfer are best depicted in Figure 3.

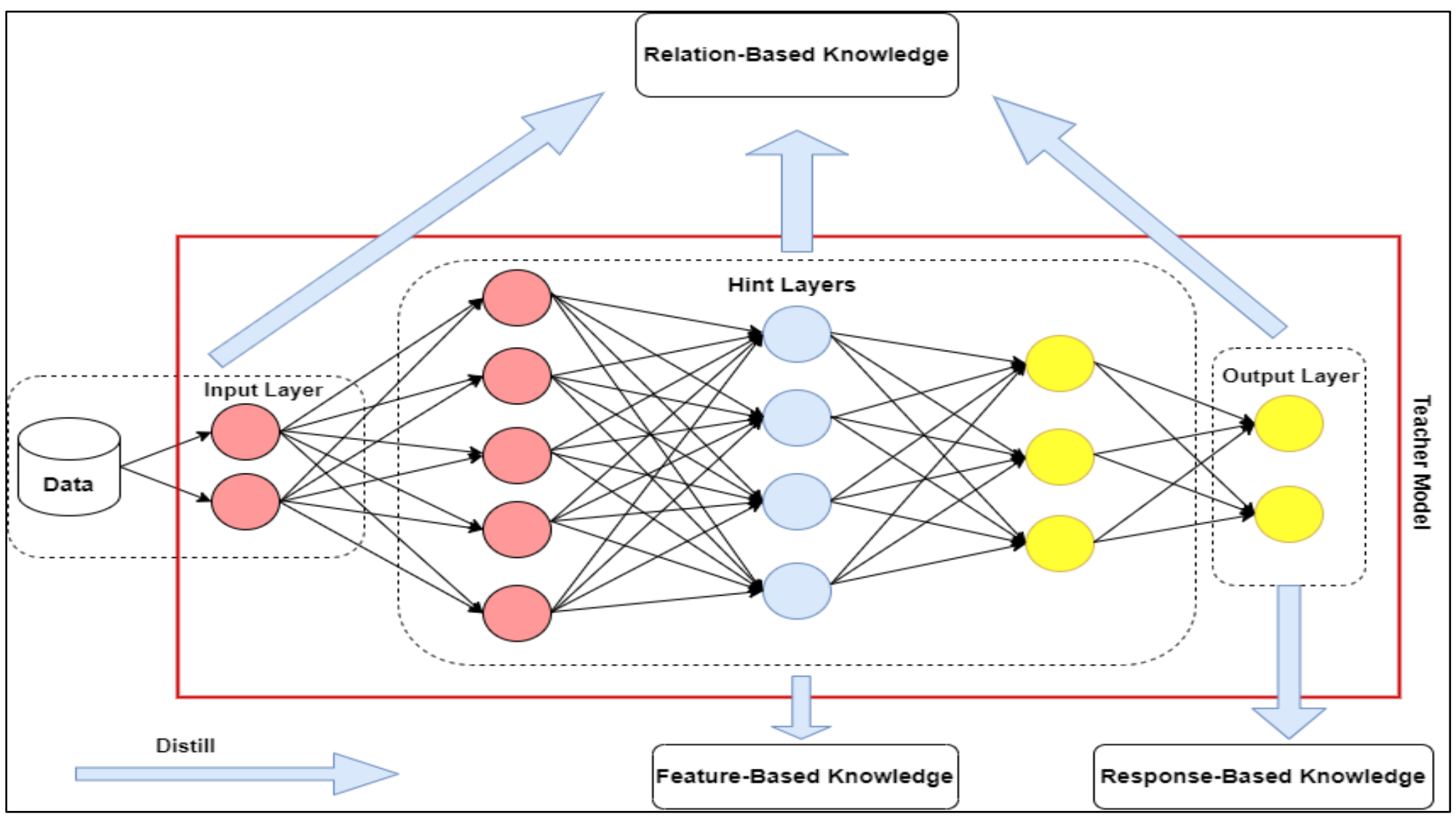

Fig 3: Response-based, Relation based, and feature-based knowledge in generic knowledge distillation architecture [53]

Knowledge distillation has been used successfully in a variety of fields, including computer vision. It has been demonstrated that it improves the performance of small models, allowing them to compete with larger and more complicated models. Among other things, knowledge distillation has been utilised for model compression, multi-task learning, and domain adaptability. The technique has become a popular research area and has the potential to enable the deployment of machine-learning models on resource-constrained devices. The authors of this [54] paper provide an extensive survey of knowledge distillation, examining knowledge categories, distillation schemes, and algorithms, as well as empirical performance comparisons. Furthermore, it examines future directions and identifies the open research challenges in KD today. Readers may refer to this comprehensive survey [54] if they would like a more detailed exploration of KD frameworks, algorithms, and related topics.

## 1.4. Distillation Schemes in Knowledge Distillation

Online and offline distillation, and self-distillation are three common distillation schemes used in knowledge distillation. Here's a brief overview of each:

### *1.4.1. Offline distillation*

Offline distillation is an increasingly popular technique in knowledge distillation for training a smaller student model to mimic the behaviors of a larger and more experienced teacher model. This approach involves training the teacher model on a huge dataset while the student model picks up the information using an offline training procedure. [55] Offline distillation provides several advantages. To start, it could reduce and simplify the model, increasing its computational efficiency and deployment speed. Second, it can enhance the student model's generalization performance, particularly when there is a lack of training data. Finally, it may be used to teach the student model the knowledge the teacher model has acquired, enhancing the student model's performance on a specific task.

Traditionally, knowledge distillation has been aimed at defining and transferring knowledge from teacher to student models effectively, but recent self-supervised contrastive methods can adversely affect the original task's representation learning. The authors in this paper [55] proposed a novel approach mitigates this by using an alternative self-supervised augmented task, which guides the network to learn the joint distribution of the original recognition task and the auxiliary task, thereby enhancing representation power without sacrificing classification capability. The use of auxiliary classifiers to intermediate feature maps generates diverse self-supervised knowledge for thorough one-to-one transfer, unlike previous methods that only use probabilistic knowledge from the final layers. The proposed approach [55] significantly outperforms previous state-of-the-art SSKD methods across various network pairs, with average improvements of 2.56% on CIFAR-100 and 0.77% on ImageNet.

In the original design, knowledge distillation was introduced as a way to transfer knowledge between teacher and student prediction logits by minimizing the KL-Divergence between teacher and student prediction logits. Distillation methods based on features outperform logit-based methods, but are computationally expensive. The classic KD loss [56] is replaced with the target classification knowledge distillation (TCKD) for predictions of target classes, and the non-target classification knowledge distillation (NCKD) for predictions of non-target classes, in order to improve logit distillation. The separation allows each part to be studied independently and adjusted accordingly. As a result, NCKD alone achieves comparable or better results than traditional KD, emphasizing the importance of non-target logits as "dark knowledge." Logit distillation's potential is limited by classical KD loss's coupled nature, which suppresses NCKD on well-predicted samples and prevents separate weighting of TCKDs and NCKDs. In order to address these issues, The authors in this paper [57] proposed Decoupled Knowledge Distillation (DKD). DKD allows independent weighting of TCKD and NCKD, thus improving distillation efficiency and effectiveness.

In spite of the fact that previous knowledge distilling methods have excelled at model compression, it is often unclear how the transferred knowledge enhances student network performance. In this study [58], the authors presented a new knowledge distillation technique that prioritizes both high interpretability and high performance. In this paper [58], they examined the mainstream CNN models, highlighting the importance of identifying regions that will be useful in classifying objects based on class. In this paper [58], they demonstrated how class activation maps can enhance this capability. This insight led them to propose class attention transfer-based knowledge distillation (CAT-KD). In contrast to prior methods, CAT-KD reveals the properties of the transferred knowledge, enhancing interpretability and understanding of CNNs. Multiple benchmarks demonstrated that CAT-KD [58], achieves state-of-the-art performance despite its clarity focus.

#### *1.4.2. Online distillation*

A knowledge distillation approach called online distillation teaches a smaller student model to mimic the actions of a larger teacher model in real-time. The teacher and student models are jointly trained on a smaller dataset during an online training procedure [59]. Updates to the student model's parameters are frequently required throughout the online distillation process in response to input from the teacher model. This can be done using a variety of techniques, including knowledge distillation loss [59], attention transfer [60], or feature imitation [61]. The objective is to improve the student model's performance on the given task while simultaneously teaching the student model to behave similarly to the teacher model.

There are various advantages to online distillation over offline distillation. First, the input from the teacher model can help direct the learning process, which can improve the convergence speed of the student model. Second, it can assist the student model in learning from cases that would otherwise be impossible to capture in a huge dataset. Finally, because the teacher model may provide instruction even when the training data is minimal, it can reduce the amount of labelled training data required for training.

While Online Knowledge Distillation (OKD) enhances models by leveraging differences between students and teachers, key issues remain unexplored, such as the impact of a large gap on student performance. In order to overcome these challenges, The authors in this paper [62] implement Switchable Online Knowledge Distillation (SwitOKD), which adjusts the distillation gap during training rather than taking into account accuracy differences during testing.

Using a switching strategy, SwitchOKD [62] switches between expert and learning modes: the teacher pauses while the student continues to learn, and the teacher resumes when the student has finished learning. As a result, an adaptive switching threshold determines when to switch modes, optimizing the distillation gap and enhancing student performance. Teachers also benefit from this method, since their performance is comparable to those obtained by other online methods. There are two basic topologies that are supported by SwitOKD: multiple networks and multi-hop networks. It has been demonstrated by the researchers [62] that SwitOKD is superior to all other methods for classification tasks based on extensive experiments.

With Online Knowledge Distillation (OKD), one can overcome the challenge of having no access to pre-trained teachers with high capacities. While current methods focus on improving ensemble prediction accuracy, they often overlook the issue of homogenization, which can cause student models to quickly saturate and underperform. The problem is thought to stem from identical branch architectures and a simplistic ensemble strategy. As a result, the authors [63] introduce AHBF-OKD, the Adaptive Hierarchy-Branch Fusion framework for OKD. Through progressively increasing branch depths, hierarchical branch architectures are designed to create diverse student peers. They [63] proposed a recursive hierarchy-branch fusion module to facilitate knowledge transfer from complex branches to simpler target branches. The training process ensures adaptive allocation of importance scores and reduces homogenization by distilling each teacher assistant from the previous hierarchy and the current hierarchy's assistant. Experimental results from the research [63] on datasets such as CIFAR-10/100 and ImageNet 2012 demonstrate AHBF-OKD's effectiveness. The distilled ResNet-18 achieves a Top-1 error rate of 29.28% on ImageNet 2012, exceeding the state-of-the-art by a significant margin.

The Knowledge Distillation (KD) online system aims to train multiple student models collaboratively, enabling them to distill knowledge from each other's experiences. Although current methods focus on class probabilities, they often overlook valuable feature representational information. For online KD, the authors [64] introduced a Mutual Contrastive Learning (MCL) framework that facilitates mutual interaction and transfer of contrastive distributions from one network to another. It [64] enhances feature representations by aggregating cross-network embedding information, which maximizes the mutual information between networks. This [64] leads to improved performance when it comes to visual recognition. Furthermore, MCL has a layer-matching mechanism that is adaptive by using meta-optimization to coordinate intermediate layers. Several experiments from this research [64] have shown that layer-wise MCL is superior to state-of-the-art online KD methods for image classification and transfer learning.

### *1.4.3. Self-distillation*

Self-distillation is a method of knowledge distillation that trains a smaller student model to act like a larger, more developed version of itself to enhance performance. After the student model has been originally trained on a dataset, its information is then compressed into a smaller student model using a self-distillation process [65]. The smaller student model is commonly trained to replicate the outputs of the bigger, more sophisticated student model by lowering the cross-entropy loss or the KL divergence between the two models' output distributions. The goal is to transfer knowledge from the bigger model to the smaller model to increase the accuracy and performance of generalization.

Self-distillation has several benefits over conventional knowledge distillation methods. First off, it doesn't call for a separate teacher model, which makes training easier and uses less computational power. Second, it may be utilised to enhance the performance of a smaller model when training data is limited. Furthermore, it may be used to iteratively enhance the smaller model, leading to performance gains. In general, self-distillation is a useful technique for improving the performance of smaller models by making use of the learnings from larger models. To provide the necessary performance improvements, it may be computationally costly and need careful hyperparameter tweaking.

The knowledge distillation (KD) method enhances generalization through the use of soft targets, but it can be time-consuming and costly to use complex pre-trained teachers and peer ensembles. Current methods of self-KD increase efficiency, but require network modifications or can be hard to parallelize. In order to overcome these issues, the authors [66] proposed a framework called Self-Distillation from Last Mini-Batch (DLB). In sequential sampling, DLB arranges each mini-batch so that half coincides with the previous iteration and the other half with the current one. With the current mini-batch, the previous iteration's soft targets are distilled, improving training stability and robustness to label noise. The implementation of DLB [66] is simple and does not require any extra memory or modifications to the model. DLB consistently outperforms state-of-the-art self-distillation methods on three classification benchmarks, and it complements augmentation strategies to enhance performance [66].
As opposed to conventional Knowledge Distillation (KD), Self-KD uses a network's own learning abilities. In this paper, the authors [67] presented Self-KD from image mixtures (MixSKD), which integrates self-distillation and mixup techniques. A mixup-generated supervisory signal helps the network learn cross-image knowledge from feature maps and probability distributions generated by the mixup. Moreover, a multi-stage feature map aggregated by a self-teacher network provides soft labels that enhance the effectiveness of self-boosting. A set of experiments from this research [67] compared MixSKD with state-of-the-art Self-KD and data augmentation methods showed its superiority in image classification, object detection, and semantic segmentation [67].

## 1.5 Role of Projector in KD

The goal of KD is to transfer knowledge from a large, powerful model to a smaller, less powerful model (student). A KD approach traditionally used for classification tasks using teacher predictions as pseudo-labels can lose important information if it is applied to non-classification tasks. It is also possible to use feature maps from earlier layers, but this poses architectural challenges and potential performance degradation. Historically, representation distillation has demonstrated success by focusing on the layer before the output space. Despite its importance in matching dimensions, this paper [68] emphasized that learned weights in the projection layer are crucial to encoding relational information. In order to achieve effective KD, it is not necessary to have complex relational structures; by using a simple linear projector with appropriate normalization schemes, essential relational information can be implicitly captured. A new theoretical perspective on distillation principles provides support for an adaptable, straightforward KD method that can be applied across various tasks and architectures [68].

As a result of complex design decisions and multiple losses between intermediate feature maps, feature distillation often results in training overhead that is unnecessary. In order to simplify this process, the authors in this paper [ 69] use a straightforward feature distillation pipeline that focuses only on features immediately before the classifier or, for generative tasks, the latent representation. The projection layer is discarded after training, so they [ 69] want to prevent it from learning representations that are not shared with the feature extractor. The present work [69] introduces a novel projection layer that maintains intra-batch feature similarity so that the representation of the students remains undistorted and knowledge transfer is maximized. In order to achieve this, it is necessary to parameterize the projection weights so that they have orthonormal rows or columns. The teachers' features can be whitening in order to enhance generality for generative tasks, as demonstrated by the improvement in performance on various distillation tasks, including image classification and object detection. The results of experiments in [69] show that the method improved state-of-the-art image classification and object detection results by up to 4.4% and 2.6%, respectively.
Although knowledge distillation has demonstrated empirical success, the current state-of-the-art methods are typically computationally expensive, making them difficult to adopt in practice. The information-theoretic framework the authors presented [70] introduces two novel, computationally cheap losses based on a cheap entropy estimator to address this issue. Students and teachers are losing correlation and mutual information in order to maximize correlation and mutual information between their representations. In addition, due to the correlation loss, the student is guided to match the traits of his or her teacher, while the mutual information loss ensures that the similarities between batches of samples are transferred. A versatile and efficient solution [70] for enhancing student networks, our approach exhibits superior performance on standard classification benchmarks, reduces the performance gap between binary and full-precision networks, and improves NLP-specific distillation methods.

## 1.6 Distillation Algorithms

Various algorithms have been proposed to improve knowledge transfer in more complex settings which are briefly discussed in Table 4.

Table 4: Summary of Distillation Algorithms

| **Distillation Algorithm** | **Description** | **Reference** |
|---|---|---|
| **Adversarial Distillation** | In adversarial distillation, a student model is taught to provide results that are like those of a teacher model while also being assaulted by an adversarial network to increase its resistance to opposed instances. | **[71]** |
| **Multi-teacher Distillation** | In multi-teacher Distillation a student model is educated using many teacher models to increase its generalizability and accuracy. | **[72]** |
| **Cross-Modal Distillation** | Cross-modal distillation is the process of transferring information from a student model that has been taught on one modality (such as text) to a teacher model that has been trained on another modality (such as images). | **[73]** |
| **Graph-Based Distillation** | Graph-based distillation involves distilling knowledge from a teacher model that has been trained on graph-structured data (e.g., social networks) to a student model. | **[74]** |
| **Attention-Based Distillation** | Attention-Based Distillation involves distilling knowledge from a teacher model that utilizes attention mechanisms (e.g., Transformer models) to a student model. | **[75]** |
| **Data-Free Distillation** | This type of KD does not require the availability of the training data used to train the teacher model. | **[76]** |
| **Quantized Distillation** | Quantized Distillation involves distilling knowledge from a high-precision teacher model to a low-precision student model. | **[77]** |
| **Lifelong Distillation** | Lifelong Distillation involves training a student model to learn continuously from multiple teacher models, each of which is specialized in a specific task or domain. | **[78]** |
| **NAS based Distillation** | NAS-Based Distillation uses Neural Architecture Search (NAS) to find a more efficient and compact student model that can still achieve high performance on a given task. | **[79]** |

## 2. Application of Knowledge Distillation in Computer Vision

The Knowledge Distillation (KD) process involves transferring knowledge from a teacher model to a student model, and is divided mainly into two categories Homogenous Architecture KD and Cross-Architecture KD. KD based on same architecture includes methods such as Attention-based KD, which guides student focus using teacher attention, and Channel-wise KD, which distills knowledge from specific channels. Additionally, Relational KD teaches feature relationships while Progressive KD progressively increases knowledge complexity, and Teacher-guided Self-Distillation guides self-distillation with predictions from the teacher. A cross-architecture KD method such as AdaMix, which dynamically selects the best teacher for each sample, and CAKD, which projects teacher features into a space that is compatible with students, are examples of cross-architecture KD methods. Multi-Teacher Knowledge Distillation incorporates predictions from multiple teachers, Feature-wise Attention Knowledge Distillation filters knowledge by using attention, and Knowledge Ensemble Distillation distills knowledge from student ensembles.
Based on advice and insights from the teacher models, these strategies seek to develop more effective student models' while preserving or enhancing performance [80–90].

Various computer vision tasks have been improved by the use of knowledge distillation, where small student models are trained to mimic large teacher models. Some of the important applications of KD in Computer Vision are discussed in detail as:

### 2.1 Image Super-Resolution

Image super-resolution models perform better without any extra training data, as Zhang et al. [91] have demonstrated. Through knowledge distillation, knowledge is transferred from a bigger teacher model to a smaller student model.

With the use of a vast quantity of high-resolution training data, a teacher model was pre-trained before being put through its paces. The second phase is to apply the information gained after the teacher model has been trained to train a student model without the need for further data. The authors provide a novel loss function that accounts for the output variation between both teacher and student models as well as the reconstruction error. As a result of this loss function, the student model can learn from the output of the teacher model while avoiding overfitting. The results of the experiments reveal that the suggested strategy outperforms existing methods that do not require additional data. For various benchmark datasets, including Set5, Set14, and B100, the approach produces cutting-edge results. Furthermore, the proposed approach depicted in Figure 4 is demonstrated to be resistant to various compression artefacts and noise levels. The research describes a viable method for improving image super-resolution models that do not require extra training data. The proposed method is straightforward and efficient as explained in Figure [91], and it has the potential to be applied to other image-related tasks that necessitate knowledge distillation. The suggested data-free knowledge distillation's conceptual framework. To synthesize images that are comparable to the original data, the generator is trained with adversarial loss and reconstruction loss. The teacher network is then gradually distilled to produce the student network.

There are situations in which the objects of interest might be small and difficult to distinguish, especially in images captured by low-definition devices and cameras. Traditional models trained on high-resolution data may have difficulty analyzing such images because of their low image resolution. In the case of low-resolution images, the application of these models directly may lead to substandard results. It may be prohibitively expensive to upgrade old low-definition cameras to higher-resolution cameras. An innovative method for classifying low-resolution images based on super-resolution guided knowledge distillation was put out by Chen et al. in 2022 [92]. In the suggested method, a teacher-student network architecture is used, with the teacher network being trained on high-resolution images and the student network being trained on low-resolution images, both of which make use of the teacher network's knowledge. A super-resolution module was created to address the problem of low-resolution images; it enhances low-resolution images before sending them to the student network. The super-resolution module is trained using a perceptual loss function and the teacher network to preserve the visual features and texture of high-resolution images. Several benchmark datasets, including CIFAR-100, CIFAR-10, and SVHN, were used to evaluate the recommended technique. The experimental findings show that the suggested technique significantly increases low- resolution image classification accuracy when compared to the most recent methodologies. For instance, the recommended method obtained an accuracy of 81.35% on the CIFAR-10 dataset while the previous state-of-the-art method only managed 78.95%. Similar results were obtained using the CIFAR-100 dataset, where the proposed method obtained 51.23% accuracy compared to the previous state-of-the-art method's 49.12%. Super-resolution guided knowledge distillation, which integrates the benefits of super-resolution with knowledge distillation, is a promising strategy for low-resolution image categorization. The suggested method improves accuracy significantly while being computationally inexpensive. Future studies might concentrate on applying this method to additional computer vision tasks and investigating its possible uses in real-world circumstances. Infrared-visible image fusion and super-resolution are two essential image processing challenges that have implications for surveillance, medical imaging, and remote sensing. These tasks are difficult because they need the integration of information from several modalities as well as the improvement of image resolution.

Recently, heterogeneous knowledge distillation has emerged as a potential approach for solving both tasks concurrently. Infrared-visible image fusion and super-resolution have been intensively researched in the literature, with numerous approaches proposed to meet these challenges. Deep learning-based algorithms have shown tremendous promise in recent years due to their capacity to automatically learn characteristics from input images. Knowledge distillation has also been frequently utilised to move knowledge from a big, complicated model to a smaller, more efficient one. The combination of these approaches, known as heterogeneous knowledge distillation, has shown promising results for infrared-visible image fusion and super-resolution. Xiao et al. [93] suggested a teacher-student network architecture for heterogeneous knowledge distillation for infrared visible image fusion and super-resolution. The teacher networks is trained on high-resolution infrared images, while the student network uses information extracted from the teacher network to train on low-resolution visible images. A fusion module is used to merge the two images, which incorporates information from both modalities. To maintain the visual features and texture of the high-resolution images, the fusion module is trained using the teacher network and a perceptual loss function. The suggested method was tested on numerous baseline data sets, including the FLIR and CAVE datasets. The results of the experiments demonstrate that the proposed method performs better in terms of infrared-visible image fusion and super-resolution than earlier methods. For instance, the recommended method provided fusion and super-resolution accuracies of 93.2% and 92.3%, respectively, using the FLIR dataset, as opposed to prior state-of-the-art methods that produced fusion and super-resolution accuracies of 90.1% and 91.4%, respectively.

### 2.2 Image Classification

### 2.3

A method for reducing knowledge in image classification issues was put out by Xu et al. [94]. The approach is based on feature normalisation, which requires a priori normalizing the characteristics of the teacher and student models. According to the authors, reducing the discrepancy between the teacher and student models can aid in advancing knowledge transmission. The properties of the teacher and student models should be normalized particularly by using batch normalisation. In the first stage of this strategy, a teacher model is pre-trained using a significant amount of high-quality training data, and then a student model is trained using the feature-normalized knowledge distillation method. The authors propose a loss function that accounts for both the classification loss and the difference in normalized features between the teacher and student models. The experimental findings suggest that the proposed technique outperforms existing non-feature normalized methods. The technique yields cutting-edge results for a variety of benchmark datasets, including CIFAR-10 and CIFAR-100. Furthermore, the suggested technique is demonstrated to be resistant to various network designs and hyperparameters. Figure 4 demonstrates the overall framework of the proposed methodology.

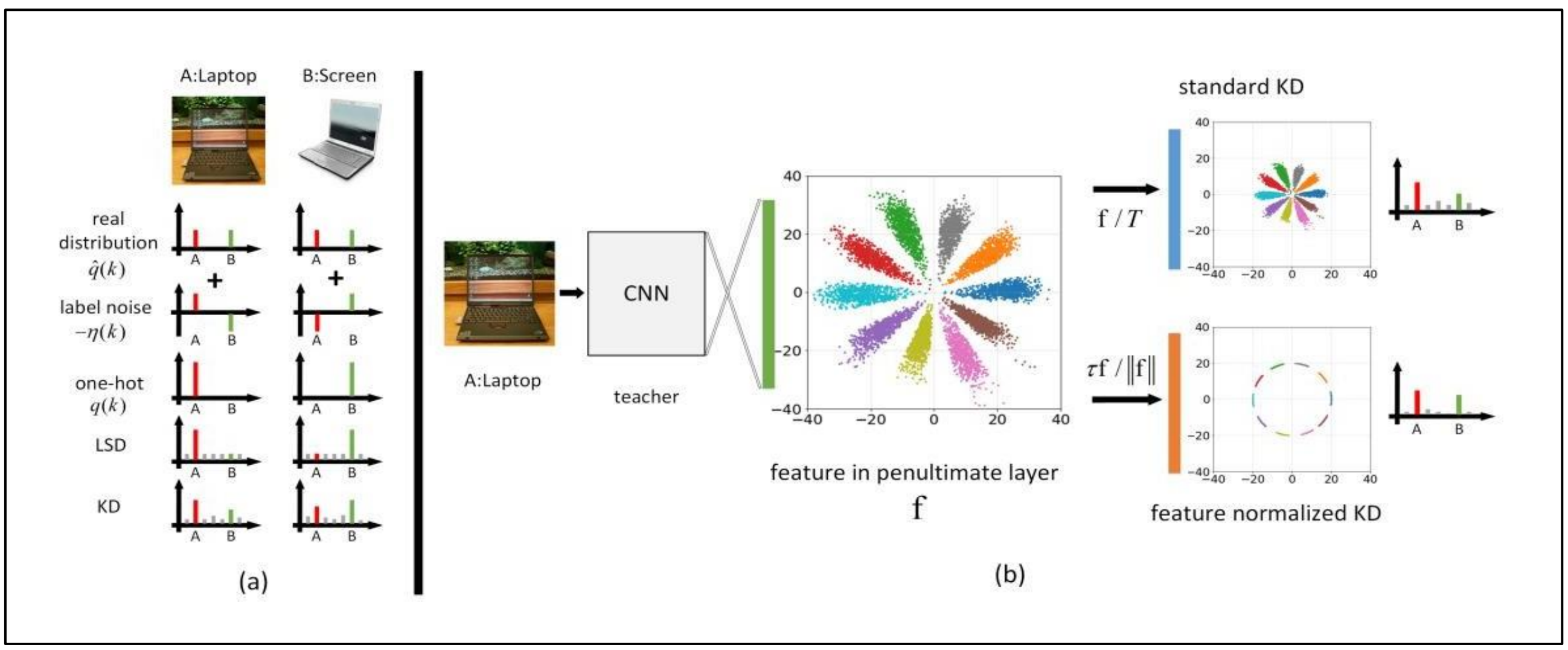

Fig 4: (a) A one-hot label that assumes classes are independent can't always properly capture an image's true distribution over classes since there are visual similarities among images. Label noise is a way of describing the distinction between a one-hot label and genuine distribution. Better supervision may be provided with LSD and KD. (b) Using our technique and a distinct kfk for each sample, KD softens the label by lowering the L2-norm of the feature in the penultimate layer with a unified T [94].

Multi-label image classification is one of the crucial tasks in computer vision. Because of the enormous number of labels involved and the intricacy of the visual information, it is challenging. Recently, knowledge distillation has emerged as a viable strategy for boosting multi-label image classification performance. Weakly-Supervised Detection (WSD) is an innovative and effective method that Liu et al. [95] introduced in 2018 to help the Multi-Label Image Classification (MLIC) task. The proposed framework (figure 5) incorporates the WSD model, knowledge distillation from WSD to MLIC, and implementation specifics. The application of a teacher-student network architecture, where the teacher network is trained on a sizable dataset of images with strong labels, while the student network is trained on the same dataset but with weak labels obtained using weakly-supervised detection.

The weak labels generated by weakly-supervised detection are less accurate than the strong labels generated by traditional supervised learning, and they just indicate the presence or absence of an object in an image without providing position information. The student network learns from the incorrect classifications and tries to predict the proper labels for the incoming images. Several datasets, including MS COCO, PASCAL VOC, and NUS-WIDE, were used to evaluate the recommended technique. The experimental results demonstrate that the proposed method outperforms existing state-of-the-art approaches in multi-label image classification by a large margin. On the MS COCO dataset, for example, the suggested technique produced a mean average accuracy (mAP) of 0.550, whereas the prior state-of-the-art method achieved a mAP of 0.533. The suggested method also performed better on the NUS-WIDE dataset, which has a huge number of labels.

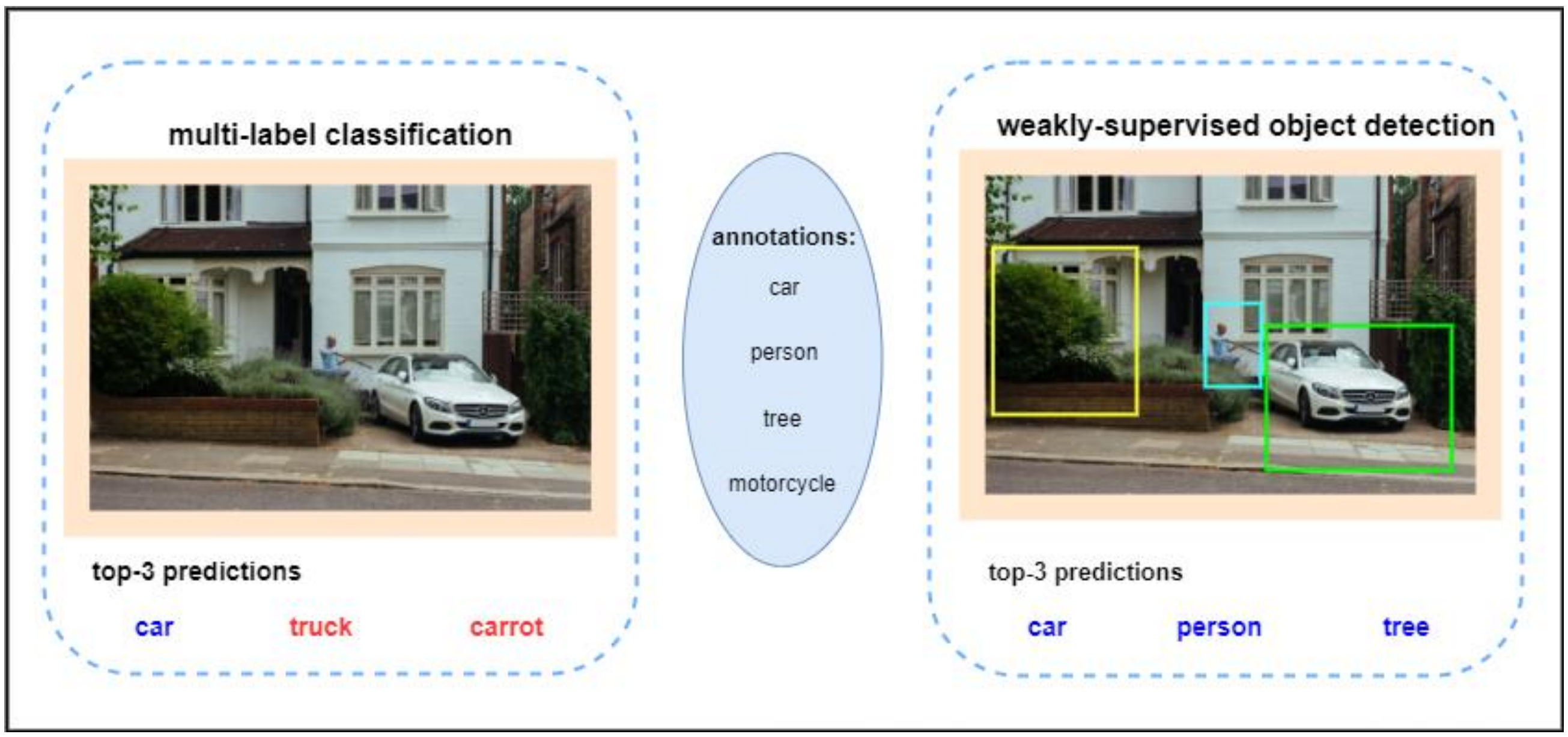

Figure 5: MLIC and WSD are depicted. Top three predictions, with correct predictions in blue and wrong predictions in red. Because of inadequate localization for semantic instances, the MLIC model may not predict effectively. Although WSD detection findings may not properly retain object boundaries, they do tend to discover semantic areas that are relevant for categorizing the target item, thus predictions can still be improved [95].

### 2.4 Medical Image Classification

CNNs have recently been used in medical image segmentation to give more precise predictions. However, the effectiveness of CNNs is heavily dependent on massive processing complexity and massive storage, both of which are unfeasible in the real world. One proposed answer to this problem is knowledge distillation. Dian et al. [96] suggested a knowledge distillation-based technique that improves the performance of medical image segmentation by applying the notion of knowledge distillation. Recognizing anatomical characteristics and anomalies in medical images, known as image segmentation, is a key difficulty in the field of medical imaging. Accurate diagnosis and treatment planning depends on proper segmentation. The suggested method is effective in reducing the computing complexity of medical image segmentation while maintaining high accuracy. Computed tomography (CT) scans, magnetic resonance imaging (MRI), and ultrasound images have all been successfully processed using this technique.

Another approach for knowledge distillation in medical image classification problems was put forth by Xing et al. [97]. Their approach is based on applying a contrastive loss function to preserve the categorical relationships between various classes. Because different medical illnesses may share traits that must be accurately differentiated by the classification model, the authors contend that maintaining categorical relations is crucial for tasks involving the categorization of medical images. The authors provide a unique contrastive loss function that encourages the teacher and student models to maintain the pairwise distances between the attributes of various classes to achieve this. A teacher model is first pre-trained using a lot of high-quality training data in the first stage of the three-stage technique that is provided. The student model is trained in a second stage using contrastive knowledge distillation that preserves categorical relations. In the third stage, the student model is modified using less labelled training data.

According to the experimental results, the proposed technique outperforms currently utilised methods that do not involve contrastive loss functions. The approach delivers cutting-edge results in a variety of benchmark medical image classification datasets, including the ChestX-ray14 and MIMIC-CXR datasets. Furthermore, it is proved that the proposed approach is resistant to different network architectures and hyperparameters. The suggested architecture is seen in Figure 6. Essentially, there are two parts to the model: the model for students and the model for mean teachers. Models for students are optimized by stochastic gradient descent, and teacher weights are updated by exponential moving averages. An image $x$ is double-augmented by two perturbations, resulting in two unique images $x_s$ and $x_t$. By using the associated augmented image as input, the student and teacher models extract feature representations $f_s$ and $f_t$ and estimate output probabilities $p_s$ and $p_t$. The weighted cross-entropy loss $L_{WCE}$ and the Kwithrgensubscript with ce $L_{KL}$ with $p_t$ oversee the student's prediction $p_s$ . To restrict the consistency of the structural information of the student and teacher, it is suggested to use $L_{CCD}$ loss, which draws positive pairings ($f_s$, $f_t$) Feature pairs with negative features are kept within the same class and pushed towards other classes with positive features.

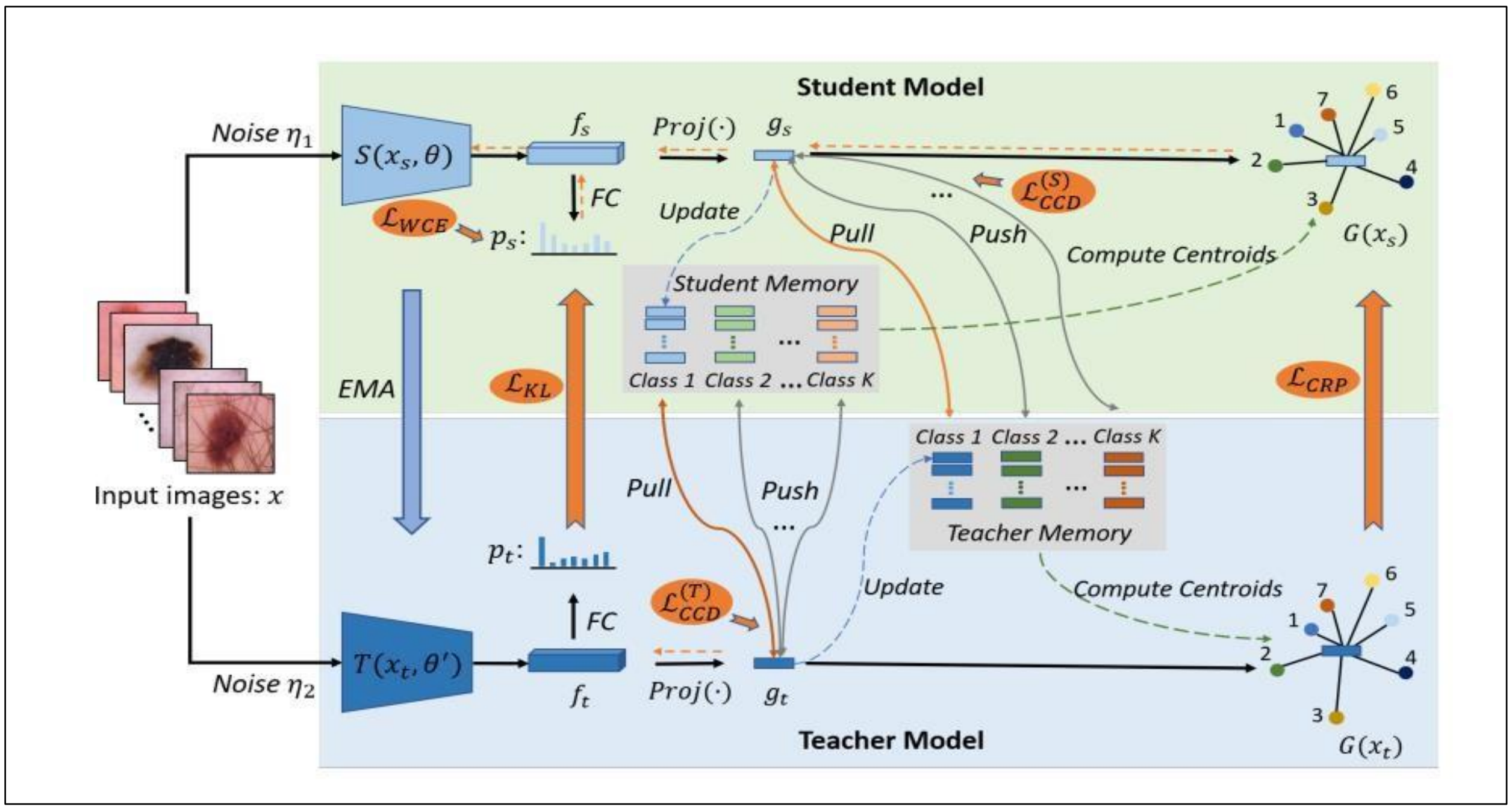

Fig 6: Categorical Relation preserving contrastive Knowledge Distillation framework [97]

### 2.5 Face Recognition

Wang et al. [98] proposed a strategy for distilling knowledge in face recognition tasks. The strategy is based on imposing exclusivity and consistency constraints on the teacher and student models' attributes. According to the authors, implementing exclusivity and consistency criteria is critical for face recognition tasks since different facial features may be associated and must be appropriately differentiated by the classification model. The authors suggest an innovative regularization term to encourage the teacher and student models to develop exclusive and consistent feature representations to do this. Pre-training a teacher model utilizing a significant amount of high-quality training data constitutes the first stage. The training of a student model utilizing the exclusivity consistency regularized knowledge distillation approach is the second stage. The authors present a new loss function that accounts for both the exclusivity consistency regularization term and the classification loss. According to the experimental results, the suggested strategy works better than other methods that do not make use of exclusivity and consistency regularization. The method generates cutting-edge results when tested against several benchmark face recognition datasets, including the LFW, AgeDB-30, and CACD datasets. The regularized and exclusive knowledge distillation process is illustrated in Figure 7. The target student network is trained using position-aware weight exclusivity and hardness-aware feature consistency.

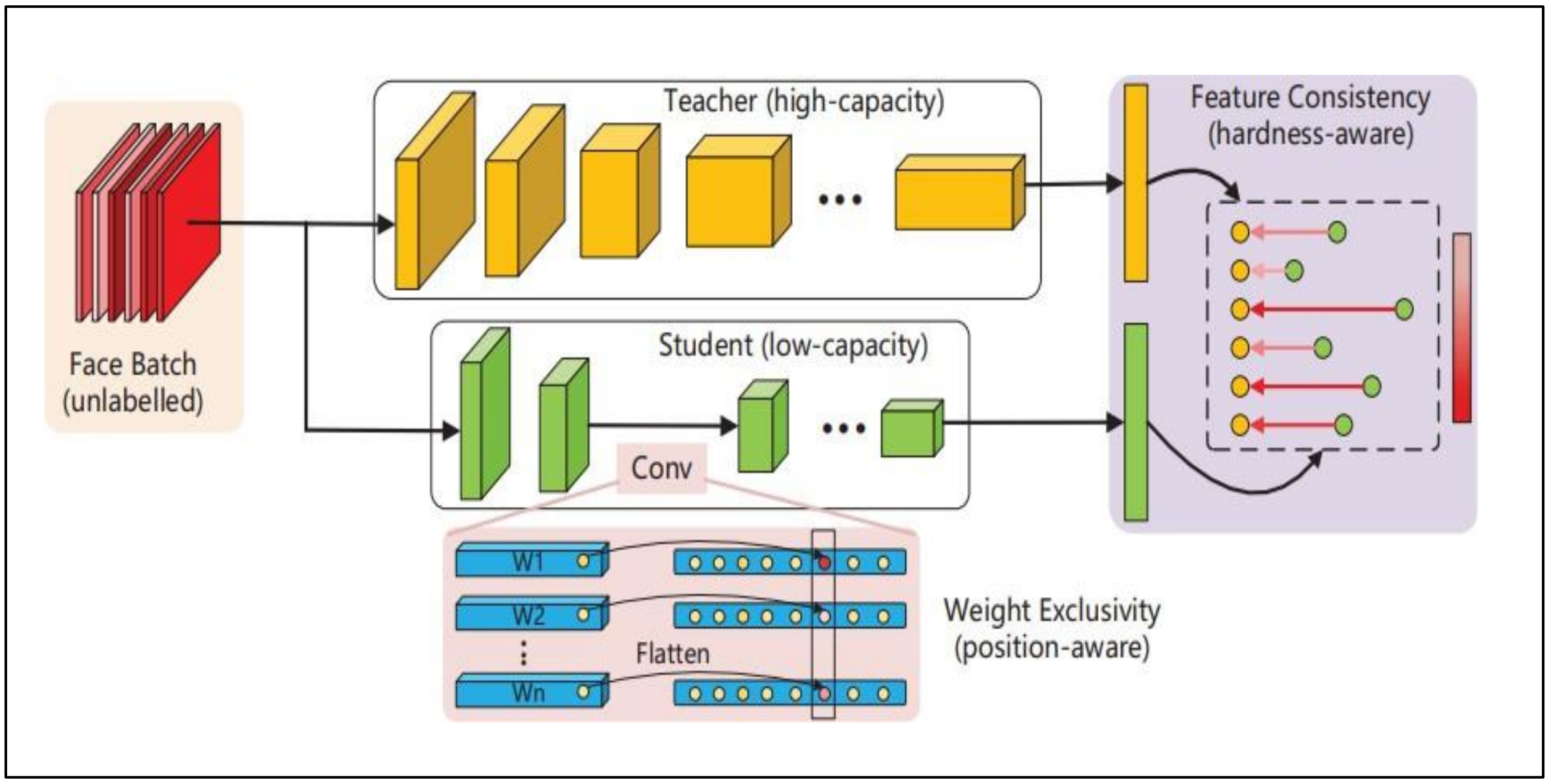

Fig 7: Knowledge distillation process that is regularized and exclusive [98].

A technique to improve the performance of face detection in low-resolution images was proposed by Ge et al. [100]. It pretrains the student network on high-resolution photos to utilised knowledge distillation to train a student network to recognizes faces in low-resolution images. Knowledge must be carefully transferred from the teacher network to the student network as part of the knowledge distillation process. Selective knowledge transfer is achieved by only passing the feature maps that are necessary for face recognition from the teacher network to the student network. The student network is optimized to recognize faces in low-resolution images and is trained to learn via selective knowledge transfer. Using selective knowledge distillation, it is possible to produce a student network that is better at recognizing faces in low-resolution images than the original teacher network. The suggested method was tested on numerous datasets, including the LFW, YTF, and IJB-A datasets, and showed a considerable improvement in face recognition accuracy in comparison with state-of-the-art approaches. The findings show that the suggested selective knowledge distillation strategy improves the accuracy of face recognition in low-resolution images. The authors also conducted ablation trials to assess the efficacy of various components of the suggested strategy. The results suggest that the selective knowledge distillation method, paired with a pre-processing phase to improve low-resolution images, yields the best results.

### 2.6 Object Detection

According to Chen et al. [101], object detection models include a large number of parameters, making them computationally costly and difficult to execute on devices with limited resources. The authors propose knowledge distillation as a solution to this problem, transferring information from a larger, more exact teacher model to a smaller, more productive student one. The teacher model is initially used to train data to create soft labels, which are probability distributions across the classes of objects in each image. The student model is then trained to distinguish items in the image while also being trained to anticipate these soft labels. When learning, the objective function combines a detection loss (which encourages the student model to correctly detect objects) and a distillation loss (which motivates the student model to produce results that are similar to those of the teacher model). The feature distillation idea, on which the distillation loss is based, calls for matching the feature maps produced by the teacher and student models. The authors suggest soft anchor distillation, a new form of feature distillation that employs a collection of anchor points to direct the distillation process. The anchor points are chosen depending on the significance of each feature map in detecting objects. The student model is utilised during inference to recognizes objects in fresh images. The authors demonstrate that the student model outperforms the teacher model in terms of memory and computation while maintaining comparable accuracy. It was observed from experimental analysis that the recommended technique is more effective than the currently used methods that do not involve knowledge distillation. For several benchmark object detection datasets, including the PASCAL VOC and COCO datasets, the approach produces state-of-the-art results. The proposed method improved the accuracy and efficiency of existing object detection models, such as Faster R-CNN and YOLO.

Another method based on the feature-based distillation involves matching the feature maps produced by the teacher and student models at different layers of the network was introduced by Linfeng et al. [102] in 2021. The authors suggest the FBKD loss, constituting a feature alignment loss and a classification loss. The feature alignment loss guarantees feature maps generated from the student model are identical to those produced by the teacher model, while the classification loss ensures that the student model can detect objects properly. The authors also present the Activation Map Selection (AMS) technique for choosing the layers utilised in feature-based distillation. For object detection, the AMS approach chooses layers depending on the discriminative ability of their activation maps. The authors first compute the mean activation map for every layer in the teacher model, which indicates the significance of each feature map in object detection. They then rank the layers based on their discriminative capability using these mean activation maps. For the AMS technique, the authors present two distinct ranking methods: global ranking and layer-wise ranking. The layers are rated in the global ranking technique based on their overall discriminative capability across the whole dataset. The layers are rated independently for each object class in the layer-wise ranking approach based on their discriminative capability for that class. The top-ranked layers are chosen for FBKD after the layers have been rated using the AMS approach. The AMS technique tries to ensure that the student model may learn the most significant properties for object recognition from the teacher model by picking the most discriminative layers.

The AMS technique tries to increase the effectiveness and precision of the student model while minimizing the memory and computation required by picking layers depending on their discriminative capability. In addition to this, we can also add new object classes to a pre-trained object detection model without retraining the entire model. This can be performed by a novel approach of class-incremental object detection which was given by Yu et al. [103]. According to the suggested architecture, the incremental learning module is in charge of expanding the student model's object classes without retraining the entire model. It achieves this by using backpropagation to change the student model's parameters. The module uses the distillation-based fine-tuning method to achieve incremental learning. The main idea is to begin learning new object classes using the information gathered from the teacher model. The module particularly instructs the student model on a small set of images made up of the new object classes, directing it using information taken from the teacher model. A soft target distribution, which is a probability distribution across the recognized object classes, is produced by the student model during training. By including a softmax function in the last layer's output of the student model, the soft target distribution is calculated. The soft target distribution is then contrasted with the target distribution, a one-hot vector that identifies the appropriate item class for each detected object. The difference between the target distribution and the soft target distribution is calculated using a loss function. The loss function is minimized through backpropagation, which modifies the student model's parameters. The student model is updated to identify the new classes when the new object classes have been refined while still remembering the prior classes that were taught using the teacher model. This technique may be performed several times to add new object classes to the student model gradually. The incremental learning module also provides a technique to prevent catastrophic forgetting, which occurs when a student model forgets previously learned knowledge while learning new object classes. The approach requires maintaining knowledge of old object classes using distillation-based fine-tuning on a limited group of images that contain both old and new object classes. This allows the student model to learn the new object classes while still remembering the old ones. The authors also give experimental findings on the Pascal VOC and MS COCO datasets that demonstrate the usefulness of the suggested technique. The results of the experiment show that the technique is successful in handling class-incremental object detection tasks, with performance equivalent to retraining the complete model.

### 2.7 Tracking Of Multiple Objects and Person Search

Due to the high processing needs of multi-object tracking and person search, Zhang et al [104] emphasize the difficulty of implementing such tasks in real-time systems. They propose utilizing knowledge distillation to overcome this difficulty, which has been successful in moving knowledge from large, more accurate models to smaller, more efficient models. Pre-training a teacher model with high-quality training data is followed by training a smaller student model using knowledge distillation. The difference in predictions between the teacher and student models, as well as classification and regression losses, are all taken into account by the authors' loss function. When predicting the item class labels for each bounding box proposal in the object detection and classification challenge, they use the cross-entropy loss as the classification loss. The ground truth bounding box coordinates and predicted bounding box coordinates are computed using the Smooth L1 loss, which is used as the regression loss. For the person re-identification challenge, they use the triplet loss as the classification loss. The triplet loss attempts to increase the distance between negative samples while decreasing the distance between positive samples in the embedding space. The Smooth L1 loss is also utilised, which determines the difference between the predicted and ground truth bounding box coordinates. The authors also present a novel loss function termed the "soft triplet loss" that is employed during the knowledge distillation process. The soft triplet loss is a variant of the triplet loss that uses soft labels rather than hard labels for positive and negative samples. It was observed from the experiments that the suggested strategy surpasses existing methods that do not involve knowledge distillation. On various benchmark datasets, including the MOT16 and MARS datasets, the suggested technique yields cutting-edge results. Furthermore, the suggested technique uses fewer computer resources while retaining excellent accuracy, making it appropriate for real-time applications.

Another paper that focuses on person search in real-world scenarios is Yaqing et al. [105]. Due to the enormous number of persons who must be tracked, occlusions, and changes in look and lighting conditions that might occur in the video, person tracking is a laborious endeavor. Furthermore, for real-time video surveillance applications, typical person search algorithms might be computationally costly and impracticable. To overcome these problems authors suggested using expert-guided knowledge distillation with a new loss function known as expert-guided attention loss (EGAL). The EGAL loss function guides the student network's attention maps using the attention maps created by the teacher network. The EGAL loss function guarantees that the student network concentrates on the same parts of the image as the teacher network, enhancing accuracy while lowering computing cost. The process of person tracking can be broken down into several stages such as; 1) Detection: In the first stage, object detection algorithms are used to identify people in each frame of the video. The detection algorithm outputs bounding boxes around each person in the frame. 2) Feature Extraction: In the second stage, bounded boxes are used to get the features. These features can include appearance-based features, motion-based features, or a combination of both. 3) Association: In the third stage, the features extracted from each bounding box in the current frame are compared with the features of previously detected bounding boxes in previous frames. The goal is to associate the bounding boxes with the same person across different frames. 4) Tracking: In the final stage, the tracking algorithm assigns a unique ID to each person and tracks their movement across different frames in the video. The suggested technique was tested on two large-scale person search datasets: CUHK-SYSU and PRW. It obtained state-of-the-art results in terms of mAP and calculation time on the CUHK-SYSU dataset. Produced mAP of 84.1%, which is greater than the previous best result of 81.4%, while lowering calculation time by 75% when compared to the teacher network. This technique also produced state-of-the-art results in terms of mAP and calculation time on the PRW dataset. The new technique produced an mAP of 62.3%, greater than the previous best result of 56.1% while lowering calculation time by 71% when compared to the teacher network.

### 2.8 Video Captioning

Video captioning is a difficult undertaking that needs specialized knowledge and a thorough comprehension of visual scenes. A method for captioning videos utilizing spatial-temporal graph convolutional networks (ST-GCN) and knowledge distillation was proposed by Pan et al. [106]. The suggested method is based on an ST-GCN architecture, a spatial temporal graph convolutional network that is made to recognizes both spatial and temporal connections in the video input. The ST-GCN is used to model the connections between several frames in a video sequence and extract features from video frames. To create captions for the video, a decoder network is fed with the features recovered from the ST-GCN. Each frame in the video is represented as a node in the Spatio-temporal network, and each edge in the graph indicates the similarity between neighboring frames. A convolutional neural network that has been pre-trained on a large-scale video classification dataset is used to determine the similarity between frames. Captions are generated via an attention-based decoder, which focuses on different areas of the video while creating each word in the caption. The decoder receives a sequence of feature vectors as input, which are created by pooling the feature maps generated by the CNN across the video frames. A loss function called temporal relation loss is used in the study Spatio

Temporal Graph for Video Captioning with Knowledge Distillation. It is employed in the video captioning challenge to impose temporal consistency in the expected sequence of words.

Let $h_i$ and $h_j$ be the hidden states of the Bi-LSTM network at time steps $i$ and $j$, respectively.

The temporal relation loss is given by:

$$L_{tr} = \sum_{\{i,j\}} max(0, \lambda - (t_j - t_i)) * ||h_i - h_j|| \quad 2^2 \tag{2}$$

*Here, $ti$ and $t_j$ are the timestamps corresponding to the $i^{th}$ and $j^{th}$ words in the caption, respectively. $\lambda$ is a hyperparameter that controls the minimum time gap between $i$ and $j$.*

To improve the performance of the suggested approach, the authors incorporate knowledge distillation into the training procedure. They use the student network's outputs as objectives while training a teacher network to offer subtitles for a specific video. Then, to improve captioning performance, the student network is taught to mimic the instructor network's actions. The student network in this scenario is a shrunk-down replica of the teacher network, which is a pre-trained spatiotemporal graph model. Distillation loss and classification loss are the two components of the loss function used to train the model. Distillation loss examines the variance between the outputs of the teacher and student models, while classification loss examines the variance between anticipated and actual captions. A weighted average of the classification and distillation losses, chosen by a hyperparameter, makes up the final objective function. The whole model is trained using stochastic gradient descent and backpropagation. The MSR-VTT and Activity Net Captions datasets are used to assess the proposed technique, and the results demonstrate that it outperforms several current solutions in terms of relevance and accuracy of captioning. To further explore the contributions of the many elements of the suggested approach and highlight the significance of each element in obtaining excellent captioning performance, the authors additionally conduct ablation experiments. Faster R-CNN object recommendations are represented by yellow boxes in Figure showing Sample of a video from MSVD with the description "A cat jumps into a box.  given in [106]. Red arrows signify directed temporal edges, whereas blue lines signify undirected spatial connections (only the most significant are depicted for clarity).

### 2.9 Label Smoothing and Knowledge Distillation

Yuan et al. [107] suggested a unified framework for combining label smoothing and knowledge distillation to improve deep neural network performance. Label smoothing is a machine learning technique used to reduce overfitting and improve a model's generalization performance. It involves substituting the actual label with a smoothed probability distribution that allocates a probability mass to each of the other classifications. This reduces the model's confidence in its predictions and drives it to learn more robust and generalized data representations. Label smoothing has been demonstrated to be useful in a variety of tasks, including image classification. Label smoothing and knowledge distillation, according to the authors, are two commonly used deep learning algorithms that can improve the accuracy of deep neural networks. However, these tactics have been studied separately, and their combined potential for improving model performance has not been fully explored. The suggested method combines cross-entropy loss and KL divergence loss into a weighted total, integrating label smoothing with knowledge distillation. The standard knowledge distillation method uses the KL divergence loss between the student model's soft predictions and the teacher model's soft goals. The KL divergence formula is as follows:

$$KL_{(P||Q)} = \sum_i P(i) \, log \, (P(i)/Q(i)) \tag{3}$$

Probability distributions $P$ and $Q$ are, respectively, over the same discrete variable $i$. In the context of knowledge distillation, P stands for the soft objectives offered by the teacher model and $Q$ stands for the soft predictions generated by the student model. The KL divergence calculates how different the two distributions are from one another. By substituting a smoothed distribution for the one-hot encoding of ground-truth labels, the authors propose label smoothing regularization. Thus, the loss function is a combination of the KL divergence loss and the cross-entropy loss between the smoothed labels and the soft predictions. The formula for the combined loss function is as follows:

$$L = (1-\alpha) * L_{CE} + \alpha * L_{KL}(P_{smooth} \,||\, Q) \quad (4)$$

where α is a hyperparameter that governs the trade-off between the two loss components and $P_{smooth}$ is the smoothened distribution of the ground-truths. Cross-entropy loss quantifies the discrepancy between the student model's soft predictions and the ground-truth labels' one-hot encoding, whereas KL divergence loss quantifies the discrepancy between the ground-truth labels' smoothed distribution and the soft prediction. The experimental findings show that the proposed framework outperforms existing approaches that only employ label smoothing or knowledge distillation. The technique achieves cutting-edge performances in a number of benchmark datasets, including CIFAR- 10, CIFAR-100, and ImageNet. Furthermore, the proposed strategy is shown to be capable of improving deep neural network robustness against adversarial attacks. The authors also provide "Teacher-Free Knowledge Distillation" (TFKD), a revolutionary approach to Knowledge Distillation. In place of using pre-trained teacher networks to provide information to student networks, TFKD proposes using the student networks themselves as teachers. This is accomplished by training an ensemble of student networks with varying initializations and then utilizing the predictions of the ensemble to regularize the training of a single student network. By doing so, the student network can learn to emulate the ensemble's behavior, improving its accuracy and generalization capacity. This method eliminates the necessity for a pre-trained teacher network and may result in improved performance in situations when a suitable teacher network is not available.

Another research by Shen et al. [108] and others examines whether label smoothing is genuinely compatible with knowledge distillation. Prior research has shown that the two methodologies are incompatible, but these studies may have been hampered by experimental design or lacked proper analysis, according to the authors. To answer this question, the authors ran extensive experiments on numerous datasets and architectures to assess the performance of models trained using label smoothing and knowledge distillation, both separately and together. They discovered that, while label smoothing can have a detrimental influence on knowledge distillation performance, it is not intrinsically incompatible with the approach. The authors specifically observed two major detrimental impacts of label smoothing on knowledge distillation. 1) Label smoothing can make the soft labels used in knowledge distillation less discriminative, reducing the student network's capacity to learn from the teacher network. This is because label smoothing replaces hard labels with a smoothed distribution, making it more difficult for the student network to discriminate between various classes. 2) Label smoothing can also result in over-regularization, in which the student network is too limited by the smoothed labels, limiting its capacity to learn from the teacher network. This can happen if the smoothing value used in label smoothing is set too high, causing the smoothed labels to be overly similar to one another. The authors offer a modified version of knowledge distillation that may be employed in conjunction with label smoothing to improve performance. The modified version of knowledge distillation proposed by the authors in the paper is called Softened Knowledge Distillation with Label Smoothing (SKD-LS).

### 2.10 Semantic Segmentation Using Knowledge Distillation

Yang et al.'s [109] approach to cross-image relational knowledge distillation was offered as a way to enhance the performance of semantic segmentation models. The main goal of the technique shown in Figure [109] is to improve segmentation precision by using the connections between different images in the collection. In addition to segmentation labels, the teacher model is explicitly trained to forecast pairwise correlations between pixels in various images. Then, on both the segmentation and pairwise connection prediction tasks, the student model is trained to replicate the teacher model's results. On several benchmark datasets, the suggested method outperforms cutting-edge segmentation algorithms in terms of accuracy. When compared to existing approaches, the recommended method also provides greater outcomes with fewer parameters and lowers computational expenses. Pixel embeddings from the same image. The symbols $ti$ and $Si$ stand for the pixel embeddings of the i-th pixel position that the teacher and student, respectively, have marked in an image given in [109]. The similarity between the two pixels is depicted by the dotted line. Using circles and lines, a relational graph is built [109].

Structured Knowledge Distillation (SKD) is a new technique for knowledge distillation in the context of semantic segmentation suggested by Liu et al. [110]. SKD works by condensing the teacher network's knowledge into a structured representation, which is then utilised to drive the student network's training. The teacher network is used to create soft labels for each pixel in the image, and the student network is taught how to foresee these soft labels using a structured loss function that promotes the student's predictions to match the teacher's predictions. The SKD procedure consists of three steps: 1) Generation of soft labels: The teacher network is used to produce soft labels for each pixel in the image. Soft labels are probability distributions across semantic classes that are more informative and resilient than hard labels. Soft labels are turned into a structured representation, which captures the spatial coherence of the semantic classes in the image. This is accomplished with the use of a graph-based representation that encodes the connections between pixels and their neighbors. 2) Structured knowledge distillation: The student network undergoes training using a structured loss function, which promotes its predictions to be consistent with both the teacher's predictions and the structured representation of the soft labels. The loss function considers the spatial coherence of the semantic classes in the image, thus assisting the student network in producing more accurate and coherent segmentations. Various types of structured loss functions can be used in semantic segmentation, including:

***Structured SVM loss***: It is a well-known loss function in object identification and image segmentation. It aims to maximize the separation between the true class and other classes while accounting for the structured interactions between pixels.

***Structured cross-entropy loss:*** Cross-entropy loss is similar to traditional cross-entropy loss, used in classification problems, this loss function takes into consideration the structured connections between pixels. It penalizes the model more severely when it misclassified neighboring pixels with different labels.

***Graph-based loss functions:*** These loss functions capture the spatial coherence of the semantic labels by using graph-based representations of the image. They usually penalize the model for providing segmentations that violate graph-based requirements like label smoothness or label consistency.

Structured knowledge distillation produced state-of-the-art results on several benchmark datasets for semantic segmentation, including PASCAL VOC, Cityscapes, and ADE20K. The authors demonstrate that their technique gets a mean intersection over union (mIoU) score of 83.6% on the PASCAL VOC dataset, which is the highest known mIoU score on this dataset at the time of publication. The authors also demonstrate that their technique outperforms other cutting-edge methods, such as knowledge distillation with attention transfer (AT) and knowledge distillation with self-attention (SAKD). It also obtains a mIoU score of 81.5% on the Cityscapes dataset, which is also the highest known mIoU score on this dataset at the time of publishing. The authors show that their technique uses many fewer parameters and computing resources than other cutting-edge methods like DeepLabV3+ and DenseASPP. This technique gets a mIoU score of 42.8% on the ADE20K dataset, which is greater than the scores produced by other cutting-edge methods such as DeepLabV3+, SAKD, and AT. Table 5 presents the summary of the research works discussed in section 4.

### 2.11 Recent multimodal models with Knowledge Distillation

Models for language-image pretraining like CLIP have gained attention for their zero-shot transfer capability and text-to-image generation capability. However, the complexity of these systems and the large number of parameters make distillation challenging, especially when dealing with the interactions between image encoders and text encoders. There remains a significant challenge in reducing the cost of CLIP distillation. The authors in this paper [111] introduce Tiny CLIP, a novel cross-modal distillation method which incorporates two key techniques: affinity mimicking and weight inheritance. With Tiny CLIP, smaller student models can successfully mimic the behavior of larger teacher models in terms of aligning visual and linguistic features. In affinity mimicking, cross-modal interactions are replicated, while in weight inheritance, pre-trained weights are transmitted from teachers to students, improving training efficiency. The Tiny CLIP process also introduces a multi-stage progressive distillation process to prevent information loss during extreme compression, making the models highly compact yet robust. A series of comprehensive experiments in [111] validate Tiny Clip’s efficacy, showing that it can halve the size of the CLIP ViT-B/32 model without sacrificing performance, and it significantly speeds up training when compared with starting from scratch [111].

Taking advantage of previous work [112-114], the authors in [115] developed advanced distillation strategies for CLIP models, with a focus on mimicry and interaction. As part of mimicry learning, students construct meaningful contrastive image-to-text relationships on the basis of features and gradients, aligning themselves with the teacher's knowledge. As a part of interactive learning, the authors [115] combine the teacher's embeddings with those of the student, so that the student learns implicitly by contrast against the teacher's embeddings, and by aggregating features at the end of training. They find that simple feature mimicry with Mean Squared Error loss performs best, while interactive contrastive learning does second best. The success of these methods [115] lies in maximizing feature similarity between teacher and student models. On MSCOCO [116] and Flickr datasets [117], this unified approach greatly improves the performance of various student architectures for zero-shot ImageNet classification and cross-modal retrieval.

Table 5: Summary of key of Knowledge Distillation works in Computer Vision

| Application | Type of Knowledge Distillation | Loss Function | Dataset | Error Rate | Key Points | Reference |
|---|---|---|---|---|---|---|
| **Image Super-Resolution** | Data-Free Knowledge Distillation | Reconstruction loss and adversarial loss | Set5 | 33.06dB (PSNR) | The research looked at the data-free compression technique for the single-image super-resolution task that is often utilized in smart cameras and smartphones. | [91], [118] |
| **Low-resolution image classification** | Super-resolution guided knowledge distillation | Cross-entropy loss and KL-divergence loss | PASCAL VOC 2007 | 22.4% | To improve the performance of a student model for super-resolving low-resolution images, it uses the principles of knowledge distillation and guided training. | [91-92], [119] |
| **Infrared-visible image fusion and super-resolution** | Heterogeneous Knowledge Distillation | Cross-entropy loss, KL-divergence loss, and Contrastive loss | BSD100 and Urban100 | 16.7% and 14.5% | In HKD, knowledge is transferred from a teacher model trained on one task to a student model to be trained on another but related task. | [91-93], [120] |
| **Image Classification** | Feature Normalized Knowledge Distillation | Cross-entropy loss and KL-divergence loss | CIFAR-100 | 17.5% | A variant of knowledge distillation, FNKD, aims to transfer knowledge by aligning the teacher and student feature representations. | [94], [121] |
| **Multi-label image classification** | Teacher-Student Distillation | Cross-entropy loss | NUS-WIDE and PASCAL VOC 2007 | 15.8% and 16.1% | Students should be able to achieve similar performance to | [94], [122] |
| | | | | | teachers while being computationally less expensive. | |
| **Medical image classification** | Attention Guided Knowledge Distillation | Cross-entropy loss | LiTS17 and KiTS19 | 17.5% and 16.7% | Incorporates attention mechanisms into the knowledge transfer process. | [95-96], [123] |
| **Medical Image Classification** | Categorical Relation Preserving Contrastive Knowledge Distillation | cross-entropy loss and contrastive loss | HAM10000 | 15.4% | Preserves categorical relations between instances during knowledge transfer. | [95-96], [124] |
| **Face Recognition** | Exclusivity Consistency Regularized Knowledge Distillation (ECRKD) | cross-entropy loss, exclusivity loss, and consistency loss. | LFW dataset | 0.97% | Combines exclusivity and consistency regularization to improve the transfer of knowledge. | [98], [125] |
| **Low-Resolution Face Recognition** | Selective Knowledge Distillation | Cross-entropy loss and KL-divergence loss. | LFW dataset | 0.89% | Focuses on selectively transferring information from the teacher model to the student model. | [98-99], [126] |
| **Object Detection** | Soft target distillation | Cross-entropy loss and KL-divergence loss. | PASCAL VOC 2007 | 35.6% | Transfers information from a teacher model to a student model using softened or probabilistic targets. | [101], [127] |
| **Object Detection** | feature-based distillation | Cross-entropy loss and KL-divergence loss. | COCO dataset | 32.8% | Transfers knowledge of feature representations from a teacher model to a student model. | [101], [128] |
| **Class-Incremental Object Detection** | Soft target distillation | Cross-entropy loss and KL-divergence loss. | PASCAL VOC 2007 | 27.2% | Combines incremental learning and soft target distillation for object detection based on class increments. | [102-103], [129] |
| **Multi-Object Tracking and** | Soft target distillation | Cross-entropy loss and KL- | MOT17 | 22.5% | Combines multi-object tracking | [104], [130] |
| **Person Search** | | divergence loss. | | | and person search with soft target distillation. | |
| **Efficient Person Search** | Expert-Guided Knowledge Distillation | Expert-guided attention loss | Market-1501 | 20.3% | Combines efficient person search with expert-guided knowledge distillation. | [104], [131] |
| **Video Captioning** | Soft target distillation | Temporal relation loss | M400 | 25.3% | Enhances the quality and performance of generated captions through soft target distillation. | [106], [132] |
| **Label Smoothing Regularization** | Soft target distillation | Cross-entropy loss and the KL divergence loss | ImageNet | 24.2% | Improves the quality of knowledge transfer during distillation by ensuring that the soft targets are not excessively confident. | [107], [133] |
| **Semantic Segmentation** | Cross-Image Relational Knowledge Distillation | Cross-entropy loss Relational loss | Cityscapes and Pascal VOC | 76.38% and 74.03% | Increases accuracy and generalization capabilities of semantic segmentation models through cross-image linkages. | [109], [134] |
| **Semantic Segmentation** | Structured Knowledge Distillation | Cross-entropy loss and Structured loss | Cityscapes dataset | 2% Improvement | Improves performance and generalization capabilities of student models. | [109], [135] |
| **Multi-Model Image Recognition** | Response based KD | Affinity mimicking Loss (CE+Distill Loss) | MSCOCO, FlickerK30 | 0.82% | Cross Model Knowledge Distillation | [136] |
| **Multi-Model Image Recognition** | Augmented Feature based distillation | KLD with CLIP distillation losses | CC3M | 0.56% | CLIP Knowledge distillation | [137] |

## 3. CONCLUSION

This review article examined the extraordinary developments of CNNs in the area of computer vision and emphasized how knowledge distillation has emerged as a potent method to overcome some of the problems that CNNs confront.

Thanks to developments in CNNs, which have become the backbone of different computer vision applications, the area of computer vision has made considerable gains in recent years. This paper started with the basics of computer vision, emphasizing its usefulness in extracting meaningful information from visual input. CNNs have transformed computer vision by learning hierarchical features from images and allowing tasks like object recognition, image classification, and semantic segmentation. In addition to CNNs, the paper presented a discussion on knowledge distillation, which is a technique for transferring knowledge from a large, expensive model (teacher model) to a smaller, more efficient model (student model). Training the student model to imitate the output probabilities of the instructor model results in increased performance and decreased model complexity.

The paper discussed in detail the architecture and components of knowledge distillation, such as soft targets, attention processes, and model compression strategies. Finally, we reviewed the applications of knowledge distillation in computer vision. Researchers have been able to overcome obstacles such as limited processing resources, deployment on edge devices, and real-time applications by exploiting knowledge distillation. Knowledge distillation has been used effectively in a variety of computer vision applications such as image classification, object identification, facial recognition, and semantic segmentation. While retaining competitive accuracy, it has exhibited gains in model size, inference speed, and resource efficiency. Although knowledge distillation has numerous benefits, the choice of teacher and student architectures, the selection of suitable loss functions, and the balancing of information transmission and model complexity are just a few of the many variables that must be carefully taken into account while designing an efficient distillation process.

**FUTURE DIRECTIONS**

- Multi-stage distillation: A big model is used to teach a smaller model in conventional knowledge distillation. Recent research however has demonstrated that employing many intermediate models between the large and small models can further increase accuracy. Investigating multi-stage distillation would be a fascinating field of study.

- Domain adaptation: Most knowledge distillation approaches assume that the large and small models are trained on the same dataset. However, this is not always the case in real-world situations. Investigating approaches for transferring information from a model trained on one domain to a model trained on another domain would be an interesting topic to study.

- Uncertainty estimation: Not only may knowledge distillation be used to compress models, but it can also be used to increase their robustness by improving their ability to estimate uncertainty. Investigating ways to incorporate uncertainty estimates into the distillation process would be an exciting field of study.

- Structured distillation: The majority of knowledge distillation strategies are concerned with compressing flat models. However, current neural network topologies frequently include many blocks with varying configurations. Investigating ways to compress these models while retaining their structural features is crucial.

- Interpretable distillation: By compressing large black-box models into smaller ones with more transparent decision-making processes, knowledge distillation may be utilized to produce more interpretable models. Exploring ways to produce more interpretable compressed models would be another future direction.